\documentclass{article}
\usepackage[utf8]{inputenc}
\usepackage{authblk}
\usepackage{setspace}
\usepackage[margin=1.25in]{geometry}
\usepackage{graphicx}
\graphicspath{ {./figures/} }
\usepackage{subcaption}
\usepackage{amsmath}
\usepackage{amsfonts}
\usepackage{algorithm}
\usepackage{algorithmicx}
\usepackage{algpseudocode}

\usepackage[style=nejm, 
citestyle=numeric-comp,
sorting=none]{biblatex}
\usepackage[colorlinks=true,
            linkcolor=blue,   
            citecolor=blue,   
            urlcolor=blue]{hyperref}

\title{SimTac: A Physics-Based Simulator for Vision-Based Tactile Sensing with Biomorphic Structures}

\author[1]{Xuyang Zhang}
\author[1,2]{Jiaqi Jiang}
\author[1]{Zhuo Chen}
\author[1]{Yongqiang Zhao}
\author[3]{Tianqi Yang}
\author[1]{Daniel Fernandes Gomes}
\author[2]{Jianan Wang}
\author[1,*]{Shan Luo}

\affil[1]{Department of Engineering, King's College London, London, U.K.}
\affil[2]{School of Aerospace Engineering, Beijing Institute of Technology, Beijing, China.}
\affil[3]{Department of Computer Science, University College London, London, U.K.}
\affil[*]{Address correspondence to: shan.luo@kcl.ac.uk}

\date{}

\begin{document}

\maketitle

%%%%%% Abstract %%%%%%
\begin{abstract}
% % Intro to Bio/tactile/simulation
Tactile sensing in biological organisms is deeply intertwined with morphological form, such as human fingers, cat paws, and elephant trunks, which enables rich and adaptive interactions through a variety of geometrically complex structures.
% % Challenges
In contrast, vision-based tactile sensors in robotics have been limited to simple planar geometries, with biomorphic designs remaining underexplored.
% % Method description
To address this gap, we present SimTac, a physics-based simulation framework for the design and validation of biomorphic tactile sensors. SimTac consists of particle-based deformation modeling, light-field rendering for photorealistic tactile image generation, and a neural network for predicting mechanical responses, enabling accurate and efficient simulation across a wide range of geometries and materials. 
% % Results
We demonstrate the versatility of SimTac by designing and validating physical sensor prototypes inspired by biological tactile structures and further demonstrate its effectiveness across multiple Sim2Real tactile tasks, including object classification, slip detection, and contact safety assessment.
% % Impact
Our framework bridges the gap between bio-inspired design and practical realisation, expanding the design space of tactile sensors and paving the way for tactile sensing systems that integrate morphology and sensing to enable robust interaction in unstructured environments. 
\end{abstract}

%%%%%% Main Text %%%%%%

\section{Introduction}
Biological organisms have evolved a wide variety of tactile sensing structures, ranging from the dexterous fingers of humans and the retractable claws of cats to the flexible tentacles of octopus and the prehensile trunks of elephants~\cite{nguyen2023bioinspiration, li2024biomimetic}. The morphological diversity of these structures is closely linked to their functional adaptability: the soft, tubular shape and sucker structures of the octopus tentacle allow it to conform to complex surfaces, while the elephant's trunk tip, with two finger-like protrusions, enables it to pinch and grasp a variety of objects. The tactile sensing capabilities of these structures further enable the perception of object properties, such as shape, texture, and stiffness, while also facilitating adaptive interactions with the environment. For instance, tactile sensing allows an organism to detect and respond to external forces, such as adjusting grip to prevent an object from slipping or rapidly contracting in response to harmful stimuli for safety~\cite{lucarotti2013synthetic}. The synergy between morphological diversity and tactile sensing in biological systems serves as a powerful source of inspiration for the design of tactile sensors. 

Tactile sensors, such as resistive~\cite{chen2019ultrahigh}, capacitive~\cite{boutry2018hierarchically}, and photoresistive~\cite{bai2020stretchable} types, are widely used in robotics for detecting contact forces and object deformations. However, conventional tactile sensors often struggle with shape adaptability and the precise capture of intricate contact details. In contrast, vision-based tactile sensors have recently emerged as a highly promising technology for achieving high-resolution tactile perception~\cite{yuan2017gelsight,donlon2018gelslim,padmanabha2020omnitact,gomes2020geltip, jiang2024rotipbot, ward2018tactip,lambeta2020digit,sun2022soft, zhang2024tacpalm, van2020large, zhang2024rotip, fan2025crystaltac, li2022implementing}. These sensors typically consist of a camera positioned underneath a soft elastomer layer, capturing visual data of the
elastomer’s deformation upon contact with an object. By analysing these deformations in captured images, a wealth of physical information can be extracted, including surface textures~\cite{luo2018vitac}, object pose~\cite{she2021cable, he2023tacmms}, shape~\cite{jiang2022robotic}, slip detection~\cite{dong2017improved}, and contact forces~\cite{chen2024transforce}. This rich data enables enhanced robotic dexterity in unstructured environments. However, existing sensors are typically limited to simple geometric shapes, such as cubic~\cite{yuan2017gelsight} and hemispherical forms~\cite{ward2018tactip}. To improve adaptability for interacting with complex surfaces, recent research has focused on developing more advanced sensor morphologies, such as finger-like structures~\cite{gomes2020geltip, sun2022soft}. These innovations not only expand the sensing area but also enhance the sensor's ability to conform to diverse object geometries, thereby enhancing robotic perception.

\begin{figure}[ht!]
\centering
\includegraphics[page=1,width=1\textwidth]{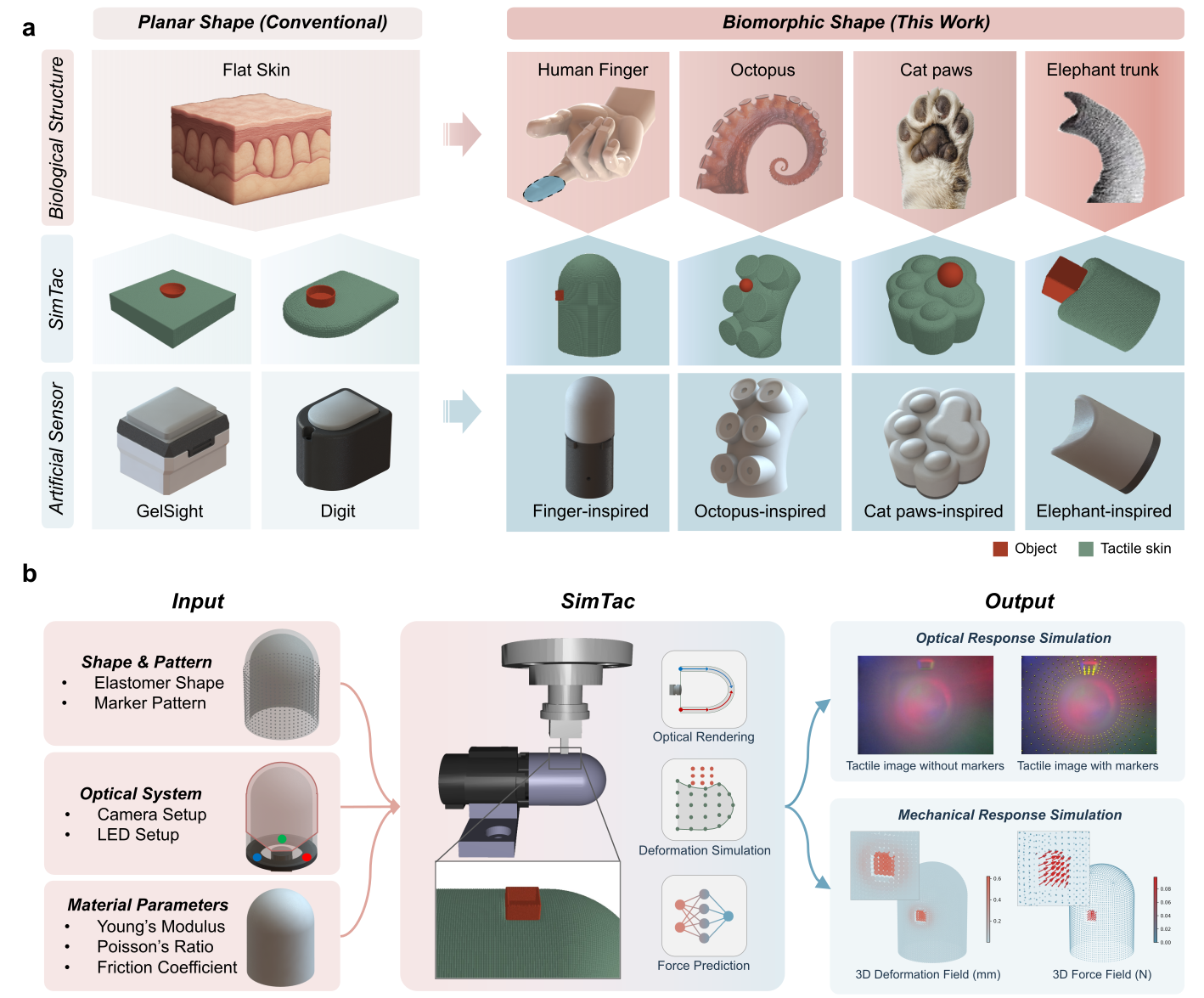}
\caption{Bridging biological and artificial tactile sensing via SimTac: a simulator for modelling biomorphic vision-based tactile sensors. \textbf{a}, Compared with conventional flat-shaped tactile sensors, biomorphically inspired artificial tactile sensors exhibit enhanced geometric complexity and an extended sensing range, which concurrently imposes increased challenges for simulation modelling. \textbf{b}, The simulation framework of SimTac. The input includes the sensor shape, marker pattern, optical system definition, and material properties. The output consists of optical responses and mechanical responses.}
\label{fig1}
\end{figure}

Designing vision-based tactile sensors with biomorphic forms through trial-and-error and iterative hardware prototyping is challenging~\cite{gomes2020geltip,sun2022soft}. Unlike simple geometric shapes, natural and organic morphologies introduce more complex skin geometries, making the deformation upon contact with objects difficult to model and estimate. Integrating key components such as cameras and LEDs within these complex geometries further complicates the design process, as it requires establishing well-controlled light paths and ensuring optimal imaging of the contact regions~\cite{gomes2020geltip,padmanabha2020omnitact}. In addition, learning-based methods for robotic perception and control typically require large-scale data collection~\cite{luo2018vitac,calandra2017feeling}. In this process, the high cost and time-consuming data collection, along with the physical wear and tear of sensors, present a major bottleneck.

% \begin{figure}[h]
%     \centering
%     \includegraphics[width=0.5\textwidth]{fig 1}
%     \caption{This is an example figure.}
%     \label{fig:1}
% \end{figure}

Simulation offers an efficient and repeatable approach to replicate physical interactions in a virtual environment~\cite{gomes2021generation,gomes2023beyond}, facilitating accelerated prototyping, systematic testing, and rapid data collection for vision-based tactile sensors. Such a simulator typically involves three key components: deformation simulation to model sensor skin behaviour, optical simulation to render tactile images, and force simulation to estimate contact forces. For deformation simulation, depth-based methods~\cite{gomes2021generation, zhao2024fots, wang2022tacto, gomes2023beyond,akinola2024tacsl, church2022tactile} approximate deformation efficiently but ignore material properties. Physics-based approaches, such as the Material Point Method (MPM) based approaches~\cite {chen2023tacchi, sun2023simulation, 10720429, sun2025tacchi, sun2025soft}, model material behaviour more accurately, but current works are limited to flat sensor geometries. Finite Element Methods (FEM) based approaches improve accuracy~\cite{10256475, cui2023tactile, du2024tacipc, si2022taxim, si2024difftactile, luu2023simulation}, yet their computational cost scales significantly with mesh density, posing challenges for real-time interactive applications. For optical simulation, data-driven methods~\cite{kim2023marker, lin2024vision, zhong2024tactgen, zhao2024fots, si2022taxim, si2024difftactile} learn the colour distribution of real tactile images for rendering but struggle with generalisation across different sensors, while physics-based rendering, e.g., Phong's model~\cite{gomes2021generation, chen2023tacchi}, struggles with complex-shaped membranes. Path-tracing-based approaches~\cite{agarwal2025vision, sun2023simulation, 10720429, 9561122, du2024tacipc} offer realism, although typically at a high computational cost. For force simulation, penalty-based methods~\cite{xu2023efficient} are efficient but less accurate than FEM-based approaches~\cite{xu2023efficient, si2024difftactile}, whose efficiency also degrades greatly with complex shapes. Overall, most existing tactile simulators are limited to flat sensor shapes, while the simulation of biomorphic sensor structures remains largely unexplored, as shown in Fig.~\ref{fig1}a. 

% structures
In this work, we present SimTac, a physics-based simulator for vision-based tactile sensors with biomorphic geometries, capable of generating accurate optical and mechanical responses in real time, as shown in Fig.~\ref{fig1}b. SimTac features three core components: a particle-based framework to simulate sensor deformation, a light field rendering system for generating high-fidelity tactile images, and a neural network for predicting dense force distributions. The simulator offers exceptional flexibility, supporting a broad range of biomorphic shapes-including human fingers, cat paws, octopus tentacles, and elephant trunks-as well as diverse optical configurations and material properties, from soft elastomers to rigid substrates. SimTac also enables zero-shot sim-to-real transfer across a variety of tactile perception tasks, such as object shape classification, slip detection, and contact safety assessment. By expanding the morphological design space for vision-based tactile sensing, SimTac opens up new possibilities for adaptive robotic systems that tightly couple morphology and sensing to interact more effectively with unstructured environments.

\section{Materials and Methods}
% The materials and methods section should provide sufficient information to allow replication of the results. This section should be broken up by subheadings. Under exceptional circumstances, when a particularly lengthy description is required, a portion of the materials and methods can be included in the Supplementary Materials. 

\subsection{System overview}
The simulator consists of three main modules: (1) A particle-based iteration model for the simulation of contact and sensor membrane deformation; (2) A light field-based lighting model for optical rendering; (3) A neural network-based model for dense force field mapping. To facilitate understanding of the proposed simulation methodology, the finger-shaped GelTip tactile sensor~\cite{gomes2020geltip} is used as the simulated target.

\begin{figure}[ht!]%
\centering
\includegraphics[width=1\textwidth]{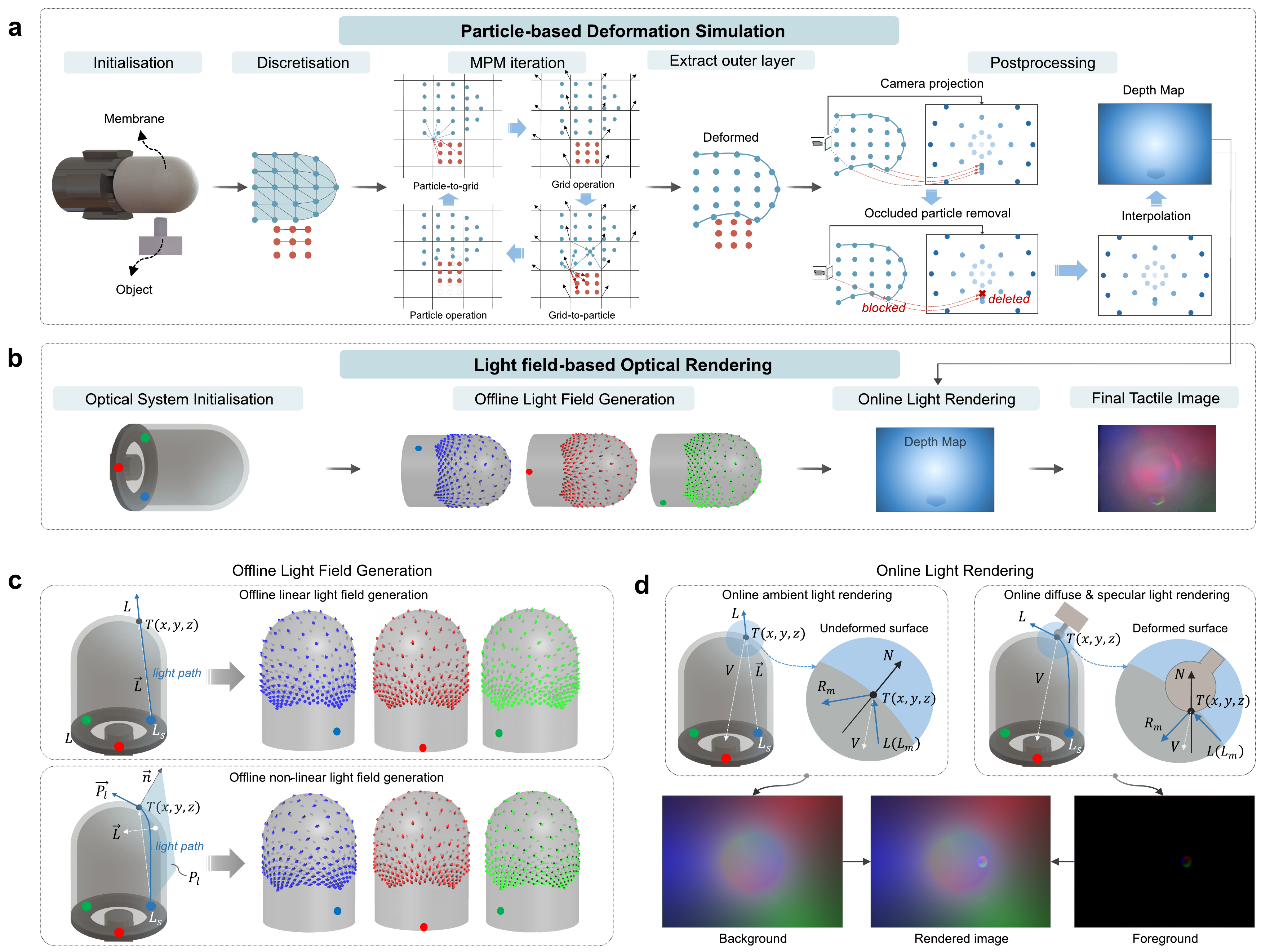} 
\caption{The principle of sensor membrane deformation simulation and optical rendering simulation. Here we use a finger-shaped GelTip sensor~\cite{gomes2020geltip} as an example. \textbf{a-b}, The pipeline of the tactile image simulation. We amplify the contact deformation and sparsify the particle density to facilitate easier understanding; a particle-based iteration method is employed for simulating membrane deformation, while optical rendering is achieved using a light field-based lighting model. \textbf{c}, Offline light field generation. \textbf{d}, Real-time online image rendering.}
\label{fig2}
\end{figure}

\subsection{Sensor membrane deformation simulation}

Fig.~\ref{fig2}a illustrates the pipeline for simulating the deformation of the sensor membrane. The process begins with the initialisation and discretisation of both the sensor membrane and the contacting object. A particle-based iterative method is then applied to compute the deformations occurring during the contact. Finally, particles are extracted and served as input for optical rendering and force prediction after post-processing.

%\vspace{1em}

\subsubsection{Initialisation and discretisation.}
We discretise the contacted object into particles using the uniform sampling method provided by \textit{Open3D}~\cite{zhou2018open3d}. For the sensor membrane model, we apply the Structured Mesh Algorithm~\cite{bern2000mesh} for uniform mesh partitioning, treating mesh nodes as particles, and recording their indices. This discretisation strategy enables uniform particleization of sensor membranes with arbitrary geometries while preserving particle indices, thereby facilitating the tracking of particles of interest, such as those located in regions associated with actuators or boundary conditions.
% We employ the Structured Mesh Algorithm~\cite{bern2000mesh} to discretize the sensor membrane model by defining the mesh type \( Element \) and node spacing \( h \). The coordinates of the generated nodes \( N \) are then extracted as the initial positions of the particles \( P \). This particle initialization method is driven by three main considerations: Firstly, it ensures uniform particle discretization of the sensor membrane, independent of its geometry. Secondly, it generates and records the indexes of all particles, facilitating the tracking of surfaces of interest, such as those related to actuators or boundary conditions. For the objects in contact, we employ a uniform sampling method from \textit{Open3D}\cite{zhou2018open3d} for particleization.

%\vspace{1em}

\subsubsection{Particle-based iteration.}
After discretising the sensor membrane and contacting objects, we employ the Material Point Method (MPM)~\cite{de2020material}, a particle-based iterative approach for simulating sensor deformation. Each particle is assigned properties such as mass, momentum, and stress, which are transferred between the particles and fixed virtual grids during each contact iteration. This process enables the update of particle velocities and displacements, allowing for the calculation of object motion and resulting deformation during interaction (see Supplementary~\ref{ST2} for details). Both the sensor membrane and the contacting objects are modelled as elastic bodies, characterised by Young's modulus \( E \) and Poisson's ratio \( v \). Rigid components are similarly treated as elastic bodies, with rigidity enforced by constraining the velocities of the relevant particles. Active motion is achieved by controlling the velocities of particles within actuator components, while boundary conditions are imposed by constraining the displacements of particles on fixed surfaces. Once the target posture is achieved (e.g., when the indenter or sensor reaches the desired position), the particle-based iteration concludes. We then extract the particles within the outer surface of the deformed membrane for post-processing.

%\vspace{1em}

\subsubsection{Data post-processing.}
% We apply different post-processing methods to the deformed particles extracted from particle-based iterations for force prediction and optical rendering. For force prediction, as shown in Fig.~\ref{fig3}a, we directly extract the initial coordinates $C_0(x_0,y_0,z_0)$ and deformation values $(\Delta{x_i},\Delta{y_i},\Delta{z_i})$ of each particle within the sensing surface. These data serve as input to the neural network for force prediction. For optical rendering, as shown in Fig.~\ref{fig2}a, the post-processing pipeline consists of three components: occluded particle removal, camera projection, and depth map interpolation. 
The post-processing pipeline for optical rendering consists of three components: occluded particle removal, camera projection, and depth map interpolation. For occluded particle removal, occlusion arises from the overlap between the foreground and background when projecting the deformed 3D particle cloud onto the 2D camera coordinate system, which is especially prominent on sensors with complex shapes or large deformations. To remove occluded particles, we utilise a ray casting algorithm~\cite{decherchi2013general} on the extracted outer surface: for each particle \( P_i(x_i, y_i, z_i) \) on the surface, we compute the ray vector \( \vec{L_i} \) from the camera position \( C_c(x_c, y_c, z_c) \) to this particle and check whether it intersects with any triangle mesh \( M_j \) within the surface. If \( \vec{L_i} \) intersects any triangle in \( M_j \) before reaching \( P_i \), the particle is considered occluded and will be removed. 

% \textit{intersects\_first} function from the \textit{Trimesh} library

For the camera projection, the deformed particle cloud $P$, with occluded particles removed, will be projected into a discrete depth map $D$ using the following camera projection model, with the camera positioned at $C_c(x_c,y_c,z_c)$:
\begin{equation}
u_i = \frac{x_i}{z_i} f_u + c_x, \quad v_i = \frac{y_i}{z_i} f_v + c_y, \quad d_i = \sqrt{(x_i - x_c)^2 + (y_i - y_c)^2 + (z_i - z_c)^2}
\label{eq1}
\end{equation}

% double check the way to calculate d

\noindent where \( (u_i, v_i) \) and \( d_i\) are the pixel coordinates and depth in the depth map \( D \), \( (x_i,y_i,z_i) \) is the coordinates of the particles within $P$, \( (c_x, c_y) \) is the centre of depth map \( D \),  \( f_u \) and \( f_v \) are the focal length components of the camera, defined as follows:
\begin{equation}
f_u = \frac{D_{\mathit{width}}}{2 \tan\left(\frac{\mathit{fov}}{2}\right)}, \quad f_v = \frac{D_{\mathit{height}}}{2 \tan\left(\frac{\mathit{fov}}{2}\right)}
\label{eq2}
\end{equation}

\noindent where $D_{width}$ and $D_{height}$ are the width and height of the depth map $D$ in pixels; $fov$ is the angular camera field of view in radians. 

For depth map interpolation, we utilise the 2D cubic spline interpolation method to interpolate the discrete depth map $D$ to obtain a smoother and continuous depth map for optical rendering. 

%Notably, due to the absence of a GPU version of the interpolation function, the frame rate of tactile image generation fluctuated around 1.5 FPS.

% The post-processing pipeline of the particles for force prediction involves directly extracting their initial coordinates $C_0(x_i, y_i, z_i)$ and deformation values $(\Delta{x_i}, \Delta{y_i}, \Delta{z_i})$ for each particle on the sensing surface. These data are then used as input to the neural network for force prediction.

%More details can be found in Appendix 4. 

%\textit{interpolate.griddata} function from the \textit{SciPy}

\subsection{Light field-based optical rendering}
% Remember to modify the text

The proposed optical rendering method includes optical system initialisation, offline light field generation and online image rendering using Phong's lighting model~\cite{phong1998illumination}, as shown in Fig.~\ref{fig2}b. For tactile sensors with biomorphic-shaped membranes, light propagation within the sensor generates highly non-linear light fields. The calculation of light propagation should not only consider the linear paths from the light source to the target points on the membrane, but also account for the non-linear light paths that follow the membrane's geometry.

%\vspace{1em}

\subsubsection{Offline light field generation.}
We generate two types of light fields: the linear light field that models light propagation along straight lines and the non-linear light field that accounts for light propagation along the curved surface of the sensor membrane, as shown in Fig.~\ref{fig2}c. 

For the linear light field, we consider the light emitted from the source light $L_s$ travelling in a straight line within the sensor to the target point $T$ on the membrane surface, as shown in the top of Fig.~\ref{fig2}c. The direction of the incident light at point $T$ can be expressed as $\vec{L} = L_s - T$. This enables us to compute the corresponding linear light field $\hat{L}_{linear}$, which includes the direction of the incident light rays at all target points $T$ located on the membrane. The collection of all target points \( T(x,y,z)\), denoted as point cloud \( \hat{P_t} \), is derived from the depth map \( D \) generated by the particle-based simulation, using the following inverse camera projection model:
\begin{equation}
x = (u - c_x)\frac{z}{f_u} , \quad y = (v - c_y)\frac{z}{f_v} , \quad z = D(u, v)
\label{eq3}
\end{equation}

For the non-linear light field $\hat{L}_{non-linear}$, light propagates through the membrane; therefore, we assume that the light path follows the membrane’s surface geometry. To compute this light path, we first define a propagation plane \( P_l \) that contains both the light source \( L_s \) and a target surface point \( T \), as shown in the bottom of Fig.~\ref{fig2}c. The vector from the light source $L_s$ to the target point $T$ is $\vec{L} = L_s - T$, while the surface normal $\vec{n}$ at point $T$ on the curved surface $z=f(x,y)$ is $\vec{n}=\left( -\frac{\partial z}{\partial x}, -\frac{\partial z}{\partial y}, 1 \right)$. The normal vector $\vec{P_l}$ of the plane $P_l$ can be calculated by using $\vec{n}$ and $\vec{L}$: $\vec{P_l} = \vec{n} \times \vec{L}$. We determine the light path by computing the intersection curve between the plane $P_l$ and the membrane mesh $M$. This is achieved by checking each triangular face within the mesh $M$ to determine whether it intersects with $P_l$ and calculating the corresponding intersection points. By connecting these points, we obtain a continuous curve that represents the intersection of the membrane mesh \( M \) with the plane $P_l$, which also corresponds to the path of light propagation within the membrane. The direction along the path at the target point $T$ is taken as the incident light direction $L$. We can compute the non-linear light field $\hat{L}_{non-linear}$ by repeating this process for all target points \( T \) on the membrane.

%More details can be found in Appendix 4. 

% \textit{intersections.mesh\_plane} function from the \textit{Trimesh} library

%\vspace{1em}

\subsubsection{Online image rendering.}
With the deformed depth map $D$ obtained from particles and the offline generated light field $\hat{L}$ from all light sources, we apply Phong's reflection model~\cite{phong1998illumination} for optical rendering, as shown in Fig.~\ref{fig2}d. The overall illumination intensity $I$ observed at a given point $T(x,y,z)$ of the sensor elastomer is given by three components: ambient, diffuse and specular light:
\begin{equation}
I = k_{a} i_{a} +\sum_{m\in \hat{L_s}}^{}(k_{d}(\hat{L}_{m}\cdot  \hat{N} ) i_{m,d}+k_{s} (\hat{R}_{m} \cdot\hat{V})^{\alpha } i_{m,s}  )
\label{eq4}
\end{equation}

\begin{equation}
 \hat{R} _{m} =2(\hat{L}_{m} \cdot \hat{N}  )\hat{N} -\hat{L_{m} } 
\label{eq5}
\end{equation}

\noindent where $\hat{L_s}$ is the set of light sources, $\hat{L}_m$ is the direction of incident light emitted from the given light source $m$ toward the given point $T$, which has already been computed in the generated light field $\hat{L}_{non-linear}$; $i_a$ is the intensity of the ambient light, which refers to the background image obtained using the linear light field rendering $\hat{L}_{linear}$ or taken from the real sensor; $i_{m,d}$ and $i_{m,s}$ are the intensities of the diffuse and specular reflections of light source $m$, respectively; $k_a$, $k_d$, $k_s$, and $\alpha$ are all reflectance properties of the surface; $\hat{R}_m$ is the direction of the reflected light; $\hat{V}$ is the direction pointing towards the camera; $\hat{N}$ is the normalized surface normals. We follow the inverse camera projection model (Eq. \ref{eq3}) to transform the deformed depth map $D$ generated from the MPM to the point cloud $P$ and compute the surface normals $\hat{N}$ using the discrete partial derivatives:

\begin{equation}
\hat{N} =\frac{\frac{\partial p}{\partial x} \times \frac{\partial p}{\partial y}}{\left \| \frac{\partial p}{\partial x} \times \frac{\partial p}{\partial y} \right \| }
\label{eq6}
\end{equation}

\noindent where the partial derivatives $\frac{\partial p}{\partial x}$ and $\frac{\partial p}{\partial y}$ are computed using the Sobel edge detector over a point $p$ in $P$. The colour of a light source is determined by its \(R\), \(G\), and \(B\) values, with the intensity of each channel calculated individually at the target point \( T \) and then combined to produce the final intensity \( (I_R, I_G, I_B) \). Notably, the methods discussed above assume the light source to be a point light source. However, other types of light sources, such as line or area light sources, can be discretised into multiple point light sources through uniform sampling~\cite{billen2016line}. The final light intensity at the target point
\( T \) is obtained by linearly summing the light intensities calculated from the light fields generated by each discrete light source.

We render the undeformed membrane using the linear light field, which serves as the background (ambient light), while the deformed area is rendered using the non-linear light field to create the foreground (diffuse and specular light). The final tactile image is obtained by overlaying the background and foreground, as shown in Fig.~\ref{fig2}d. Notably, for existing tactile sensors, we can use the tactile image collected from the real undeformed sensor as the background, whereas for new, non-existent tactile sensors, the image rendered using the linear light field will be utilised as the background.

\subsection{Force prediction model}
\begin{figure}[ht!]%
\centering
\includegraphics[width=1\textwidth]{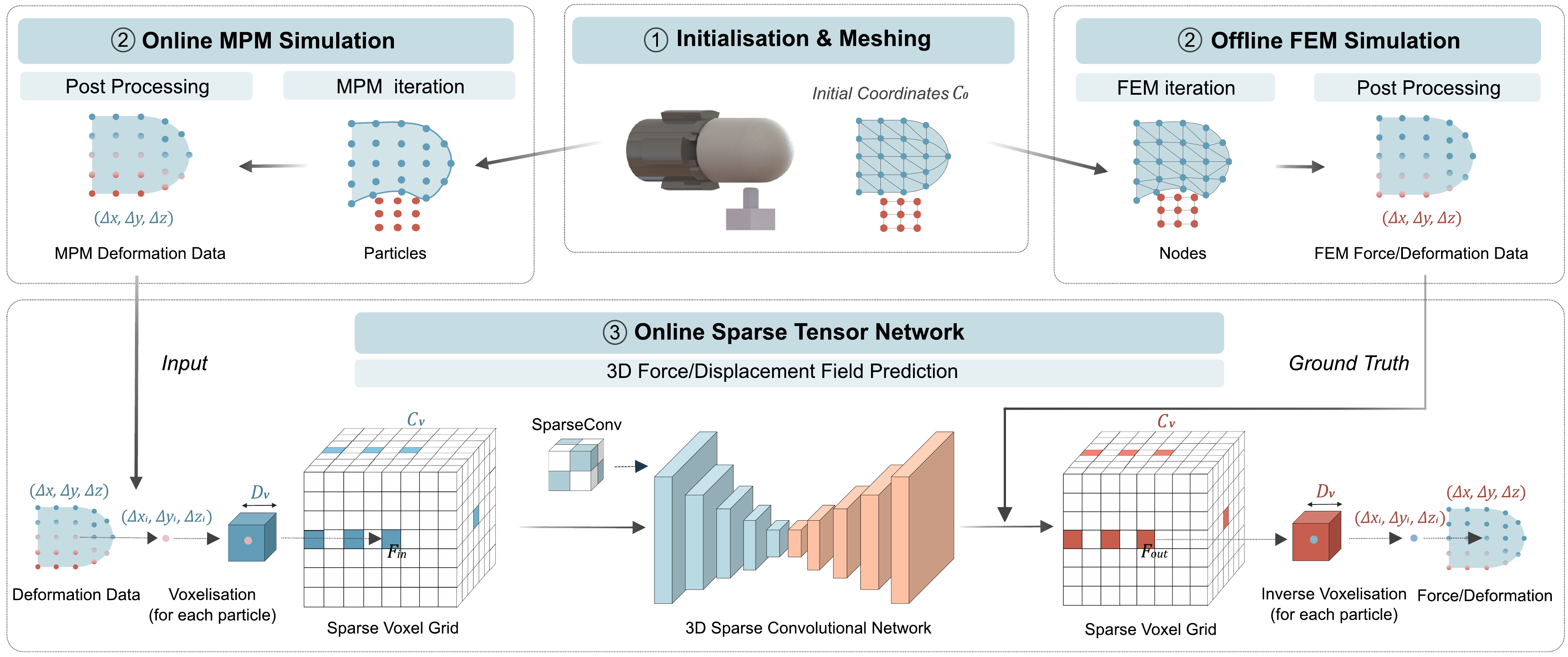}
\caption{The principle of mechanical response simulation. We amplify the contact deformation and sparsify the particle density to facilitate easier understanding; A Sparse Tensor Network is used to predict force/deformation fields, where the deformation data iterated from the MPM simulation approach serves as input, while the force/deformation data computed from FEM serves as the ground truth. This framework enables rapid and accurate mechanical response simulations that approximate FEM-level precision.}
\label{fig3}
\end{figure}

Fig.~\ref{fig3} illustrates the simulation pipeline for tactile mechanical response. After initialisation and discretisation, the deformation data of the sensor membrane obtained from MPM iterations are post-processed and fed into the Sparse tensor networks (STN)~\cite{liu2015sparse, choy20194d} to predict force. FEM simulation data are generated offline in advance and serve as the ground truth for training the STN. Notably, in addition to force distributions, other field data, such as deformation and stress, can also serve as network outputs for training and prediction.

% The post-processing involves directly extracting the initial coordinates $C_0(x_i, y_i, z_i)$ and deformation values $(\Delta{x_i}, \Delta{y_i}, \Delta{z_i})$ for each particle on the sensing surface.

Sparse tensor networks are advanced techniques for processing high-dimensional sparse data. Their core principle is to sparsely encode multidimensional correlations within tensors, effectively reducing storage and computational costs while preserving key features between neighbouring particles, making it particularly well-suited for processing particle cloud data. By leveraging STN, the sensor deformations generated by MPM can be mapped to the force/deformation distributions computed by FEM, enabling fast and accurate simulation of mechanical responses that approximate FEM-level precision.

A sparse tensor consists of a voxel coordinate matrix $C_v$ and an associated feature matrix $F$. $C_v$ is obtained by voxelising a particle cloud with a predefined voxel size $D_v$, while $F$ is determined by aggregating the particle feature vectors within each voxel. The voxel size needs to be reasonably set to prevent multiple particles from falling into the same voxel, thereby preserving the particle cloud features as accurately as possible. Therefore, we first generate the initial coordinates $C_0(x_i,y_i,z_i)$ of the particles through meshing the sensor membrane, with the membrane in its undeformed state, and extract the field data of each particle obtained from MPM and FEM iterations. The initial particle coordinates $C_0$ are then voxelized with a predefined voxel size $D_v$ to obtain the voxel coordinates $C_v$. The field data - particle displacements $(\Delta{x_i},\Delta{y_i},\Delta{z_i})$ computed by the MPM are set as input features, and the feature matrix $F_{in}$ is determined by averaging the feature values of all particles within the same voxel.

\begin{equation}
C_v = \begin{bmatrix}
b^{1}  & c_{x}^{1} & c_{y}^{1} & c_{z}^{1}\\
\vdots & \vdots & \vdots & \vdots\\
b^{N} & c_{x}^{N} & c_{y}^{N} & c_{z}^{N}\\
\end{bmatrix}, \\
F_{in} = \begin{bmatrix}
f_{x}^{1} & f_{y}^{1} & f_{z}^{1}\\
\vdots &\vdots &\vdots\\
f_{x}^{N} & f_{y}^{N} & f_{z}^{N}
\end{bmatrix}
\label{eq7}
\end{equation}

\noindent where $b^{i}$ is the batch index for particle $i$, $\{c_x^i, c_y^i, c_z^i\} \in \mathbb{Z}^3$ are the voxelised coordinates, and $\{f_x^i, f_y^i, f_z^i\} \in \mathbb{R}^3$ represent the features. The index $i \in [1, N]$, where $N$ is the number of voxels after voxelisation, which depends on the voxel size $D_v$.

The sparse convolution takes a sparse tensor as input and also produces a sparse tensor as output. In this case, the input and output share the same coordinate matrix, meaning $C_v^{in} = C_v^{out}$. The feature vector $\mathbf{f}^{\text{out}}$ for an output coordinate $c$ is then computed using the following formula:

\begin{equation}
\mathbf{f}_{\mathbf{c}}^{out} =\sum_{s\in \mathcal{N}(c,K) }^{} W_{s} \mathbf{f}_{c+s}^{in}, \quad  \mathbf{f}_{c}^{out}\in F^{out} , \quad \mathbf{c}\in C_v^{out}
\label{eq8}
\end{equation}

\noindent where $s$ represents the offset used to locate the corresponding input coordinates within the $c$-centred neighbourhood covered by the kernel size $K$, denoted as $\mathcal{N}(c, K)$. $f_{c+\mathbf{s}}^{\text{in}}$ denotes the input feature vector at the input coordinate $c + \mathbf{s}$, while $W_{\mathbf{s}}$ represents the coefficient learned during the training process. Using the coordinate matrix $C_v^{\text{out}}$ and the feature matrix $\mathbf{F}^{\text{out}}$, the output sparse tensor can be generated.

The bottom of Fig.~\ref{fig3} illustrates the overall structure of the proposed network. We first voxelise the particle cloud and convert the particle displacements into a sparse tensor. The network is based on the Minkowski UNet14 architecture~\cite{choy20194d}, which follows a symmetric encoder-decoder structure. Both the input and output consist of three-channel field data corresponding to the X, Y, and Z directions. The encoder extracts features from the sparse tensor, progressively reducing the number of feature vectors while increasing the channel dimensions. Conversely, the decoder decreases the channel dimensions while restoring the number of feature vectors. After passing through the output layer, the feature vectors have a shape of \(N \times 3\), where \(N\) represents the number of voxels, and \(3\) denotes the feature dimensionality. Finally, we perform inverse voxelisation on the output sparse tensor to obtain the particle force/deformation.

The network is trained using the L1 loss (MAE) function and optimised with the Adam optimiser. The dataset is split into training, validation, and test sets in a 3:1:1 ratio. Training is performed with a batch size of 32, an initial learning rate of \(1 \times 10^{-3}\), and a learning rate decay factor of 10 every 10 epochs. The final model is saved after 100 epochs of training.

\subsection{Experimental Setup}

We evaluate SimTac across four key aspects:
\begin{itemize}
    \item \textbf{Accuracy:} We evaluate the performance of SimTac by comparing the simulation results with ground truth data. Quantitative accuracy is assessed under different conditions, including contact with objects of various sizes, shapes, and textures, as well as different contact positions, postures, and motions.
    \item \textbf{Efficiency:} We evaluate the runtime performance of each module within the simulator to evaluate its computational efficiency, which is vital for tactile-based real-time simulations that require extensive iterations.
    \item \textbf{Flexibility:} We evaluate the flexibility of SimTac by simulating tactile sensors with different shapes and material parameters, which is crucial for simulating biomorphic-shaped tactile sensors, rather than being limited to sensors with fixed shapes and materials.
    \item \textbf{Applicability:} We evaluate the applicability of SimTac from two main perspectives: sensor prototyping and tactile-based Sim2Real tasks. For sensor prototyping, we first design sensor prototypes within SimTac, then fabricate the corresponding real-world sensors, using an elephant trunk-shaped sensor as a representative example. For perception tasks, we assess SimTac’s performance across several tactile-based Sim2Real tasks, including object classification, slip detection, and contact safety assessment, using a finger-shaped sensor as the test case.
\end{itemize}

\begin{figure}[ht!]%
\centering
\includegraphics[width=0.95\textwidth]{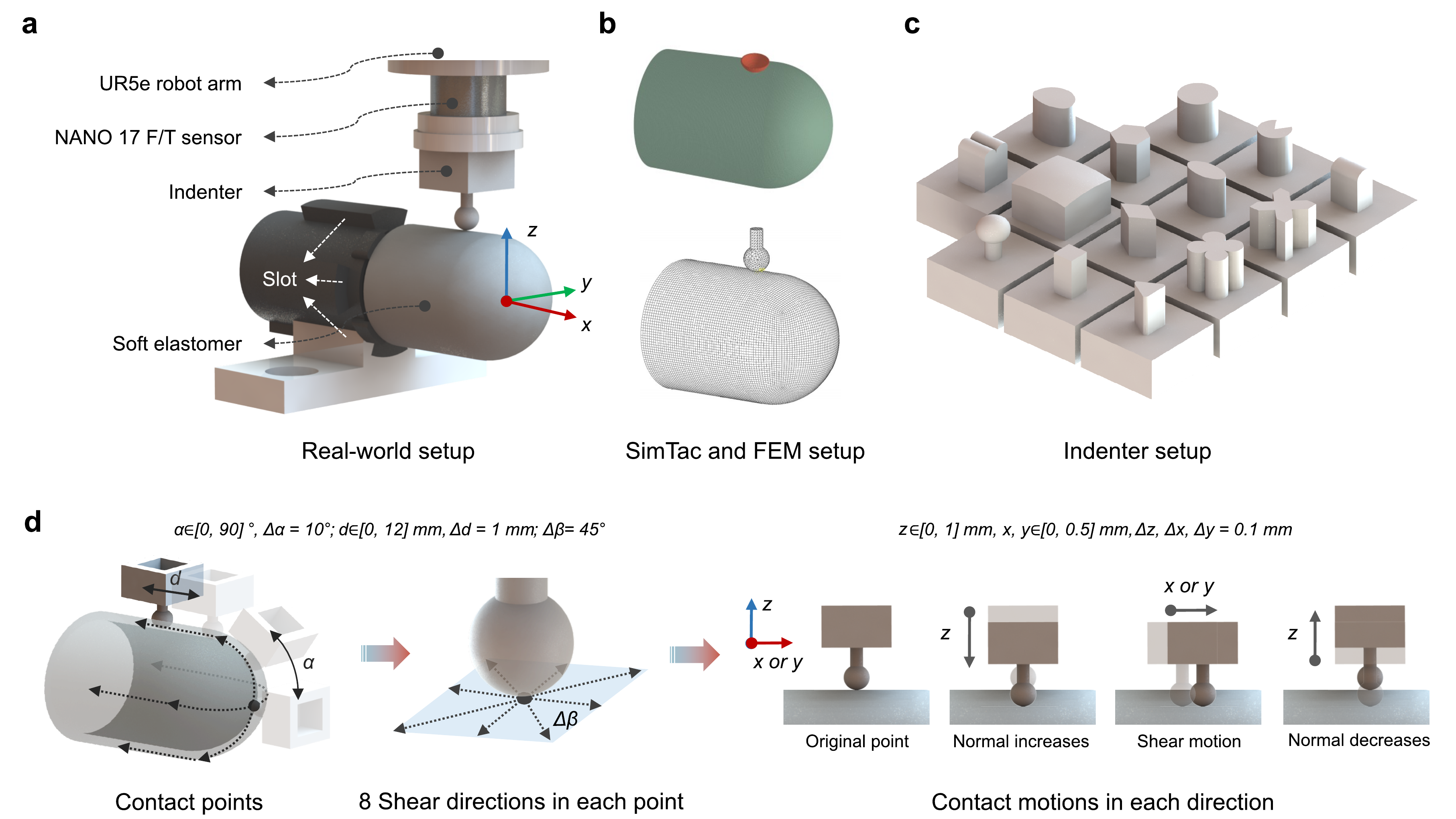}
\caption{Data collection setup on the finger-shaped GelTip sensor. \textbf{a}, Real-world data collection platform, capable of synchronously capturing tactile images and contact forces, with the indenter's contact posture controlled by the robotic arm. \textbf{b}, Configuration in SimTac and FEM replicated from the real-world scenario. \textbf{c}, Indenters with different shapes used in the experiment. \textbf{d}, The movement trajectory of the indenter during the data collection process. The contact points cover most of the sensor's surface, including a combination of vertical and tangential motion.}
\label{fig4}
\end{figure}

We employed the finger-like tactile sensor GelTip~\cite{gomes2020geltip} for the evaluation of accuracy and efficiency. GelTip closely resembles a human finger in size and shape, featuring high curvature and an omnidirectional sensing surface. Compared to flat sensors, it exhibits more complex deformations and internal light fields, making the simulation more challenging. Meanwhile, compared to other biomorphic shapes, its more uniform and regular geometry simplifies the design of data collection trajectories for the corresponding real sensor.
% Indenter Setup & collection setup & contact pose & sensor setup
Two types of GelTip sensors was configured: one without markers, used to collect tactile images for optical response comparison, and the other with markers on the skin (where the motion of the markers indicates the deformation of the sensor skin), used to collect images with marker motions and total contact force for mechanical response comparison (refer to Fig.~\ref{fig1}b, shape \& pattern input, markers have been added to the sensor membrane). To collect tactile data with the GelTip sensor, we secured the GelTip in place and mounted the indenter along with the ATI NANO 17 force/torque sensor onto a UR5e robotic arm. We controlled the robotic arm to facilitate contacts between the indenter and the GelTip at various positions and orientations. Tactile data was collected using 14 different indenter shapes, covering different contact positions, depths, and shear motions. Further details can be found in Fig.~\ref{fig4} and Supplementary~\ref{ST1}.

\section{Results}
\subsection{Accuracy evaluation}
This evaluation focuses on two key aspects: optical response simulation performance and mechanical response simulation performance. For the evaluation of optical response, we use tactile images collected from real-world sensors as references. For the evaluation of mechanical response, we adopt two reference sources: dense deformation and force fields generated by FEM, as well as tactile images with sparse marker motion and total force measurements collected from real sensors. 

\begin{figure}[ht!]%
\centering
\includegraphics[width=1\textwidth]{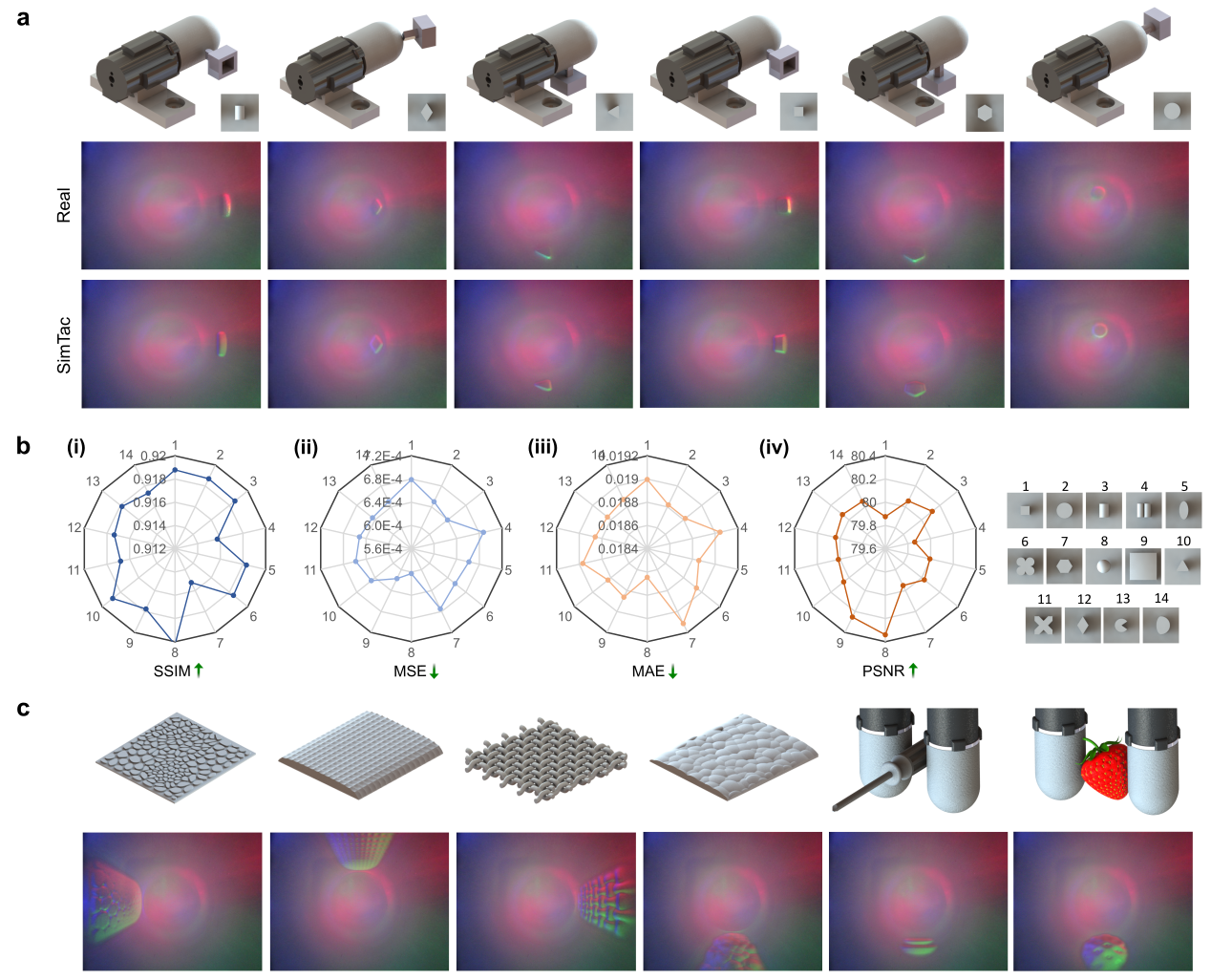}
\caption{The performance evaluation of optical response simulation. \textbf{a}, The comparison of tactile images collected from the real world and those generated from the proposed simulator under varying contact positions and indenter shapes. \textbf{b}, The quantitative analysis results of the optical simulation under contact with different indenters, evaluated using: (\textbf{i}) Structural Similarity Index (SSIM); (\textbf{ii}) Mean Squared Error (MSE); (\textbf{iii}) Mean Absolute Error (MAE); (\textbf{iv}) Peak Signal-to-Noise Ratio (PSNR). \textbf{c}, The optical simulation results when the sensor interacts with different textures.}
\label{fig5}
\end{figure}

%\vspace{1em}
% 1.2 Optical simulation performance
\subsubsection{Optical response simulation performance.}
% principle/reason and setup
% MPM + LIGHT FIELD
% To achieve optical simulation for biomorphic-shaped tactile sensors, we employ a particle-based simulation approach to iterate sensor deformation during contact and use a light field-based lighting model for rendering. 
% The particle-based method represents objects with particles and introduces a virtual grid for contact simulation, effectively avoiding the mesh distortion issues of FEM~\cite{lee1993effects} while benefiting from GPU acceleration. The light field-based lighting model enables the generation of light propagation paths tailored to different sensor membrane shapes. 
% As shown in Fig.~\ref{fig2}a, we begin by applying a mesh partitioning method to uniformly discretise the sensor model into particles. After several iterations of the deformation simulation, we project the surface particles of the sensor into the camera coordinate system, remove occluded points, and interpolate to generate a depth map, which is then used for rendering (details can be seen in~\nameref{Method}). 
We followed the experimental setup and data collection methods outlined in Fig.~\ref{fig4} and Supplementary~\ref{ST1}, collecting 6,538 data pairs in both the simulation and real-world environments to evaluate the accuracy of the simulator.
% results and analysis 
% indenter/texture/YCB object/BKG image
Fig.~\ref{fig5}a compares the real tactile images with their corresponding simulated images for various indenter shapes, contact positions, and orientations. The results demonstrate that the simulated images accurately reproduce the shape distortion caused by the deformation area being close to the wide-angle lens, and successfully simulate the deformation of the contact area, the colour distribution of reflected light, and the formation of specular highlights. To quantitatively assess the similarity between simulated and real tactile images, we employ four widely used image similarity metrics: Structural Similarity Index (SSIM)~\cite{hore2010image}, Mean Squared Error (MSE), Mean Average Error (MAE), and Peak Signal-to-Noise Ratio (PSNR)~\cite{hore2010image}. Fig.~\ref{fig5}b presents the distribution of these metrics, where the simulated tactile images show high similarity and low pixel error compared to the corresponding real tactile images across all indenter shapes and contact poses (More details can be seen in Supplementary Fig.~\ref{fig12}). Moreover, our simulator is capable of optically simulating fine surface textures and real object contacts. Fig.~\ref{fig5}c showcases textured surfaces of various types, along with objects from the YCB dataset~\cite{calli2015ycb}, which feature more intricate contact details and larger object sizes compared to the indenters.

% Additionally, we also compare the background images generated by SimTac without deformation with real-world background images, as shown in Fig.~\ref{fig2}e. 

%\vspace{1em}
% 1.3 Motion and force map simulation performance
\subsubsection{Mechanical response simulation performance.}
\begin{figure}[ht!]%
\centering
\includegraphics[width=1\textwidth]{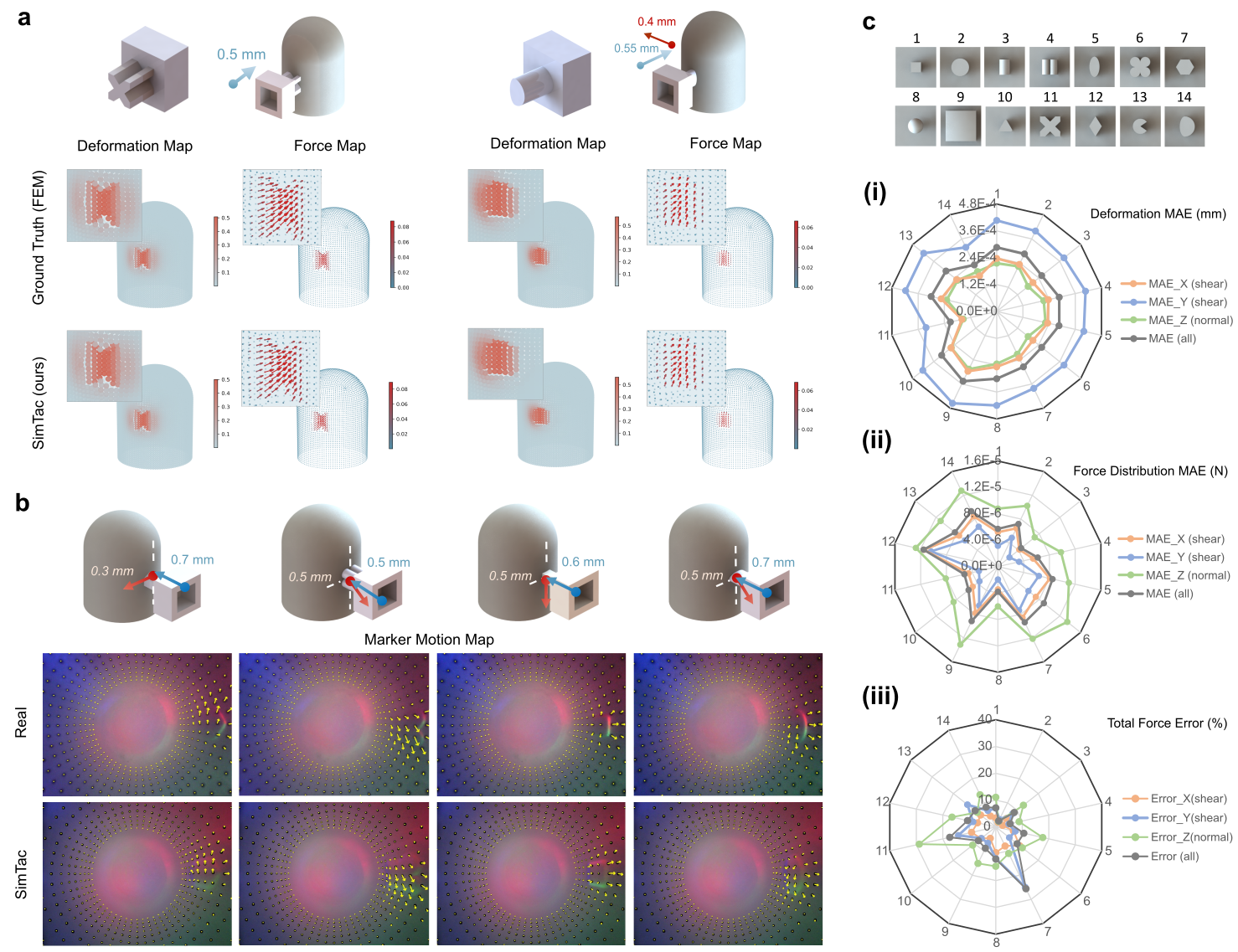}
\caption{The performance evaluation of mechanical response simulation. \textbf{a}, Comparison of 3D deformation and force fields between SimTac simulation and ground truth from FEM on the test set across different contact scenarios. \textbf{b}, Comparison of marker motion between the SimTac simulation and ground truth from the real world across different contact scenarios. \textbf{c}, The quantitative analysis results of the mechanical response simulation when in contact with different indenters. (\textbf{i}) MAE of deformation field; (\textbf{ii}) MAE of force field; (\textbf{iii}) Percentage error of the total force.}
\label{fig6}
\end{figure}

% principle/reason and setup
% results and analysis 
Following the experimental setup and data collection methods outlined in Fig.~\ref{fig4} and Supplementary~\ref{ST1}, we collect 9,980 pairs of tactile data using both the particle-based simulation approach and FEM, with data from 10 seen objects used for model training (Fig.~\ref{fig6}c, indenter 1-10), and data from 4 unseen objects used for testing (Fig.~\ref{fig6}c, indenter 11-14).  Accuracy is evaluated on the dataset collected from all indenters. For the dense reference data from FEM, we performed direct comparison, while for the sparse reference data from reality, we converted the dense fields into sparse fields through downsampling deformation field data or summing force field data to calculate the total force before comparison.
% describe how we evaluate (seen/unseen, FEM/Real， MAE/R2, how we calculate them) /the principle in simple version(model， input, output, seen/unseen)/describe the evaluation results/analysis 

We first evaluate the dense field results, as shown in Fig.~\ref{fig6}a, heatmaps are utilised to visualise the comparison between the deformation/force fields generated from SimTac and the ground truth computed via FEM on unseen indenters (see more results in Supplementary Fig.~\ref{fig13}). The results demonstrate that the neural network achieves highly accurate particle-level predictions of dense deformation/force fields in both normal and tangential directions. The simulated field data exhibit excellent smoothness and continuity, driven by the neural network’s ability to capture nonlinear relationships between adjacent particle features and effectively extract key local characteristics. To further quantify the accuracy, we compute the MAE between the SimTac-generated fields and the FEM-computed fields across the entire dataset, as shown in Fig.~\ref{fig6}c(i) and (ii). SimTac achieves high accuracy in simulating dense deformation and force field, with test set MAEs of \(2.77 \times 10^{-4}\) mm (deformation) and \(8.6 \times 10^{-6}\) N (force), averaged over the X, Y, and Z directions. Across the entire dataset, the MAEs are \(2.84 \times 10^{-4}\) mm and \(7.4 \times 10^{-6}\) N, demonstrating strong predictive performance and generalisation to diverse object shapes and poses in sensor contact scenarios (A detailed quantitative analysis can be found in Supplementary Fig.~\ref{fig14}).

To evaluate the sparse displacement field, we downsample the simulated dense deformation field to match the sparsity of the markers from the real sensor. The downsampled field is then projected into the camera coordinate system to obtain the corresponding marker positions. These markers are overlaid onto the rendered tactile images, as shown in Fig.~\ref{fig6}b. This figure illustrates the comparison between simulated and real-world results for various indenters under different contact conditions, where the yellow arrows indicate the direction and magnitude of the marker motions. To evaluate the sparse force data, we aggregate the simulated force field along the X, Y, and Z axes to compute the total force. The accuracy is assessed by calculating the MAE between the simulated and real-world force measurements, shown in Fig.~\ref{fig6}d(iii). The MAE between the predicted total force and the real one across the entire dataset is 0.021 N (13.18\% of the actual force) in the X direction (shear direction), 0.013 N (9.24\%) in the Y direction (shear direction), and 0.134 N (6.27\%) in the Z direction (normal direction). Notably, the normal force exhibits a much larger magnitude compared to the shear force, resulting in correspondingly larger prediction errors.

\subsection{Efficiency of SimTac}
% time cost of each module/FEM, data collection
We evaluated the efficiency of SimTac by measuring the online time consumption of each module, including sensor deformation iteration, optical response simulation, and mechanical response simulation, all deployed to simulate the finger-like tactile sensor GelTip and run on the GPU. All simulations were conducted on an Ubuntu 20.04 system with an i7-13700HX 16-core processor and an NVIDIA GeForce RTX 4060 GPU. The performance of the particle-based deformation simulation was evaluated under three different input particle quantities (40K, 300K, and 1.3M), resulting in frame rates of 250 frames per second (FPS), 33 FPS, and 10 FPS, respectively. The optical rendering performance was assessed at three output image resolutions (320$\times$240, 640$\times$480, and 1280$\times$960), yielding frame rates of 100 FPS, 25 FPS, and 10 FPS, respectively. The neural network-based force/deformation prediction efficiency was tested with three point cloud sizes (1K, 5K, and 25K), achieving frame rates of 100 FPS, 76 FPS, and 62 FPS, respectively.

% 2. Generalisation Capability
\subsection{Flexibility of SimTac simulator}

% material + principle
% sensor shape/illumination setup 

For vision-based tactile sensors, the optical response is primarily influenced by the sensor shape, camera parameters, and internal light sources, while the mechanical response is mainly affected by the material properties of the sensor membrane. To evaluate the flexibility of the proposed simulator, we first validate its optical response on tactile sensors with different optical setups, especially for those tactile sensors with biomorphic shapes, and then verify its mechanical response with varying material parameters.

\begin{figure}[ht!]%
\centering
\includegraphics[width=1\textwidth]{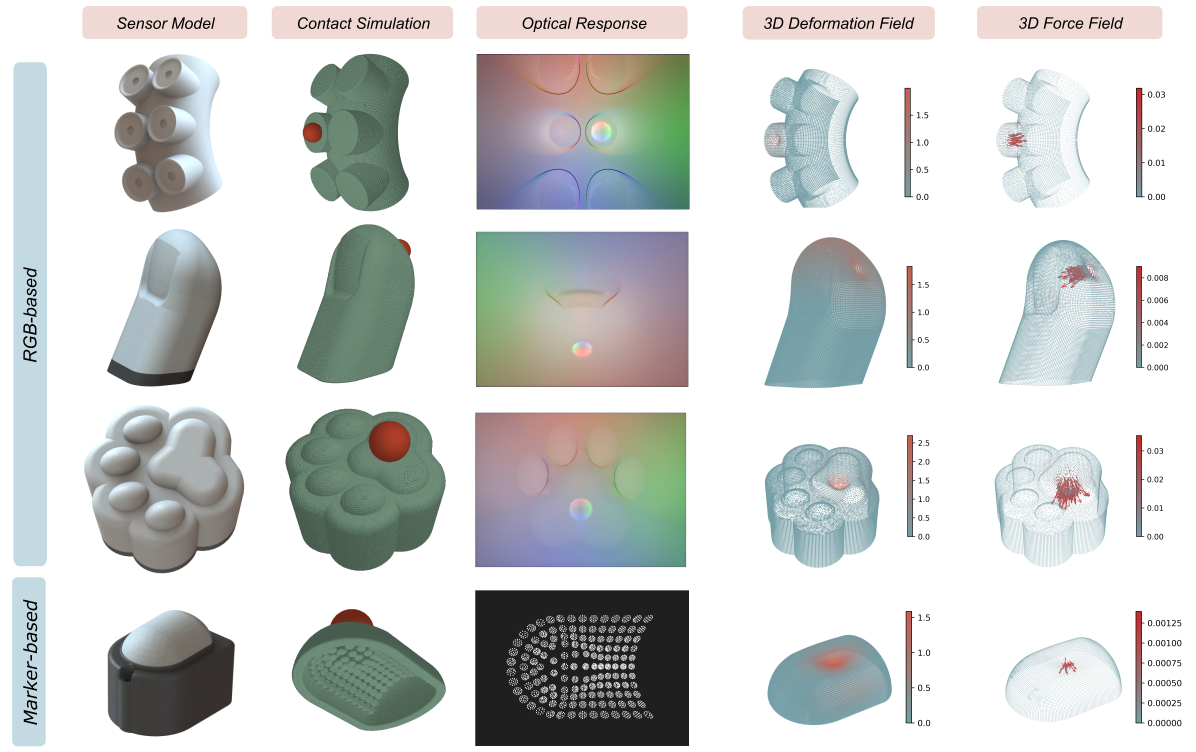}
\caption{The flexibility of the SimTac simulator. The proposed simulator can be applied to tactile sensors with diverse biomorphic shapes, and is also capable of simulating both RGB-based and marker-based tactile sensors.}
\label{fig7}
\end{figure}

To evaluate the flexibility of SimTac across different sensor configurations, we designed tactile sensors with diverse shapes, sizes and optical setups, including variations in LED placements and quantities, as well as camera positions and lens angles. Specifically, our study features four types of tactile sensors: three novel biomorphic tactile sensors modelled after the shape of an octopus tentacle, a human thumb and a cat’s paw, and a curved, marker-based DigiTac sensor~\cite{lepora2022digitac}. Fig.~\ref{fig7} shows the simulation results of tactile feedback during contact with a spherical object, including contact deformation, optical response, and mechanical response (with further details in Supplementary Fig.~\ref{fig15}). Our approach can be applied to the simulation of biomorphic-shaped tactile sensors, offering potential applications in the development of biomorphic soft robots. Moreover, for the marker-based DigiTac sensor, our method can simulate the motion of physical pins~\cite{lepora2022digitac} during contact.

\begin{figure}[ht!]%
\centering
\includegraphics[width=1\textwidth]{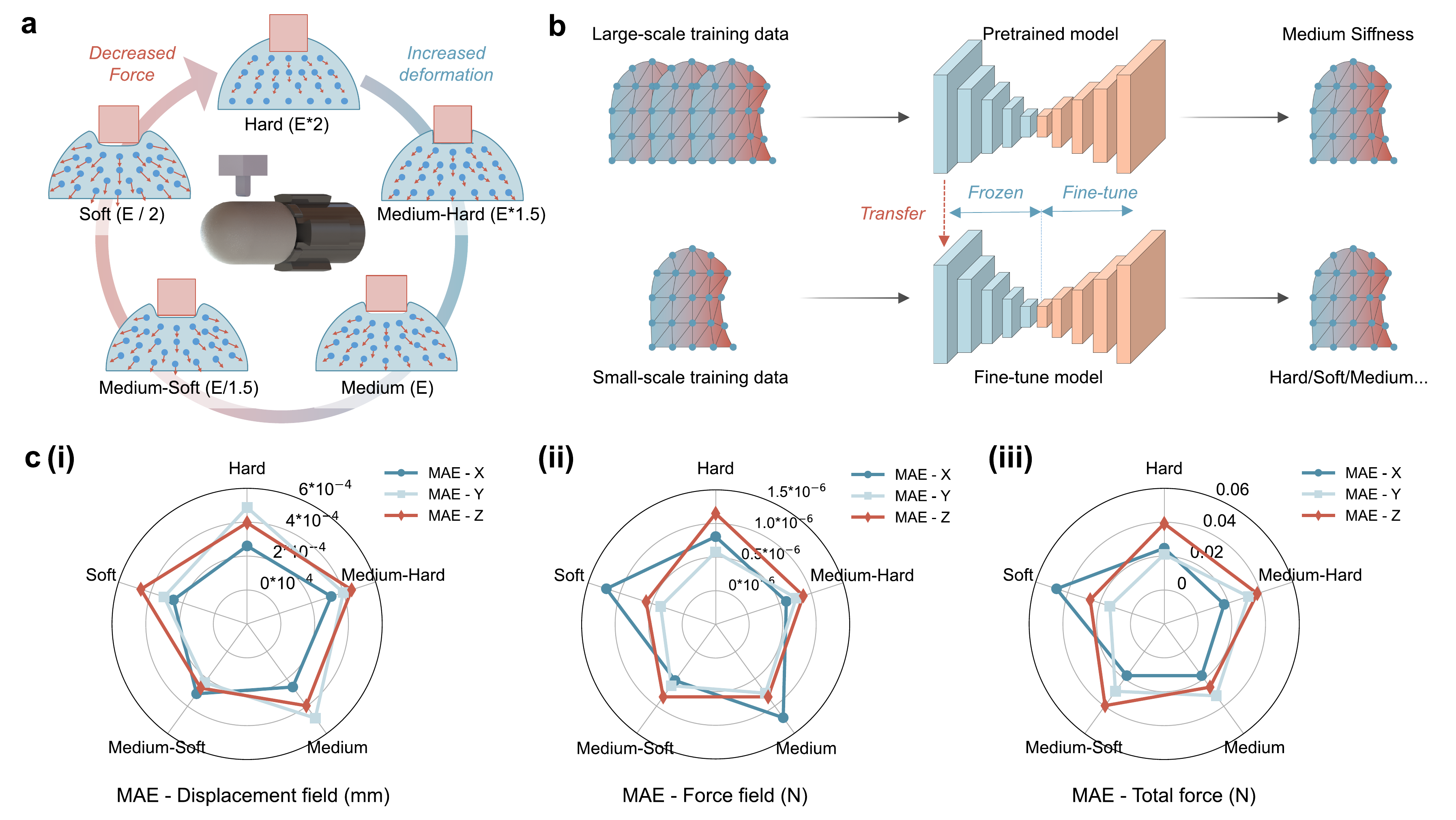}
\caption{The flexibility of the SimTac simulator. \textbf{a}, Deformation of membranes with varying hardness under identical contact conditions. Softer membranes exhibit larger deformation and lower force. \textbf{b}, The trained network can be applied to sensor membranes of different materials through a fine-tuning process, requiring only a small additional FEM ground truth dataset. \textbf{c}, The MAE of (\textbf{i}) deformation field; (\textbf{ii}) force field and (\textbf{iii}) total force in the X, Y, and Z directions between the simulated and ground truth values, evaluated across GelTip sensor membranes of different materials.}
\label{fig8}
\end{figure}

To evaluate the flexibility of SimTac on sensor skins with different material properties, we modified the original Young modulus of the sensor membrane (0.145 MPa - Medium), doubling it (0.29 MPa - Hard) to create a stiffer contact surface and halving it (0.0725 MPa - Soft) to create a softer one (Fig.~\ref{fig8}a). For each stiffness variation, a limited set of ground truth data is collected from FEM simulations to fine-tune the pre-trained neural network (trained using data with Medium stiffness), where only the decoder is updated while the encoder remains fixed (Fig.~\ref{fig8}b). We then evaluate the fine-tuned neural network on membranes of corresponding stiffness, using MAE to measure the deviation between simulated results and ground truth. As visualized in the radar chart in Fig.~\ref{fig8}c, the MAE for the deformation field remains below \(4.37 \times 10^{-4}\) mm, the MAE for the force field stays under \(1.16 \times 10^{-6}\) N, and the MAE for total force is also within 0.042 N across membranes ranging from soft to stiff. These results demonstrate that the mechanical response simulation method can be applied effectively to sensors with different material parameters through the fine-tuning strategy.

\subsection{Application of SimTac}
We first demonstrate SimTac's sensor prototyping capability by designing a biomorphic-shaped tactile sensor in simulation, with an elephant-trunk shape as an example, and then transferring the design to the real world to fabricate the corresponding physical sensors. We further evaluated SimTac's applicability in three tactile-guided Sim2Real tasks using a human finger-shaped tactile sensor: object classification, slip detection, and contact safety assessment. These tasks reflect key aspects of biological tactile perception, enabling the interpretation of object properties, monitoring of dynamic interactions, and avoidance of harmful contact. For the object classification task, the simulated optical response was utilised to infer the shapes of contacted objects. In the slip detection task, both the simulated optical responses and mechanical responses were employed to determine whether an object was slipping relative to the sensor surface. For contact safety assessment, the simulated force fields were used to evaluate the risk of damaging the sensor skin.

% %\vspace{1em}

\subsubsection{Simulation and fabrication of an elephant trunk-shaped tactile sensor.}
\begin{figure}[ht!]%
\centering
\includegraphics[width=1\textwidth]{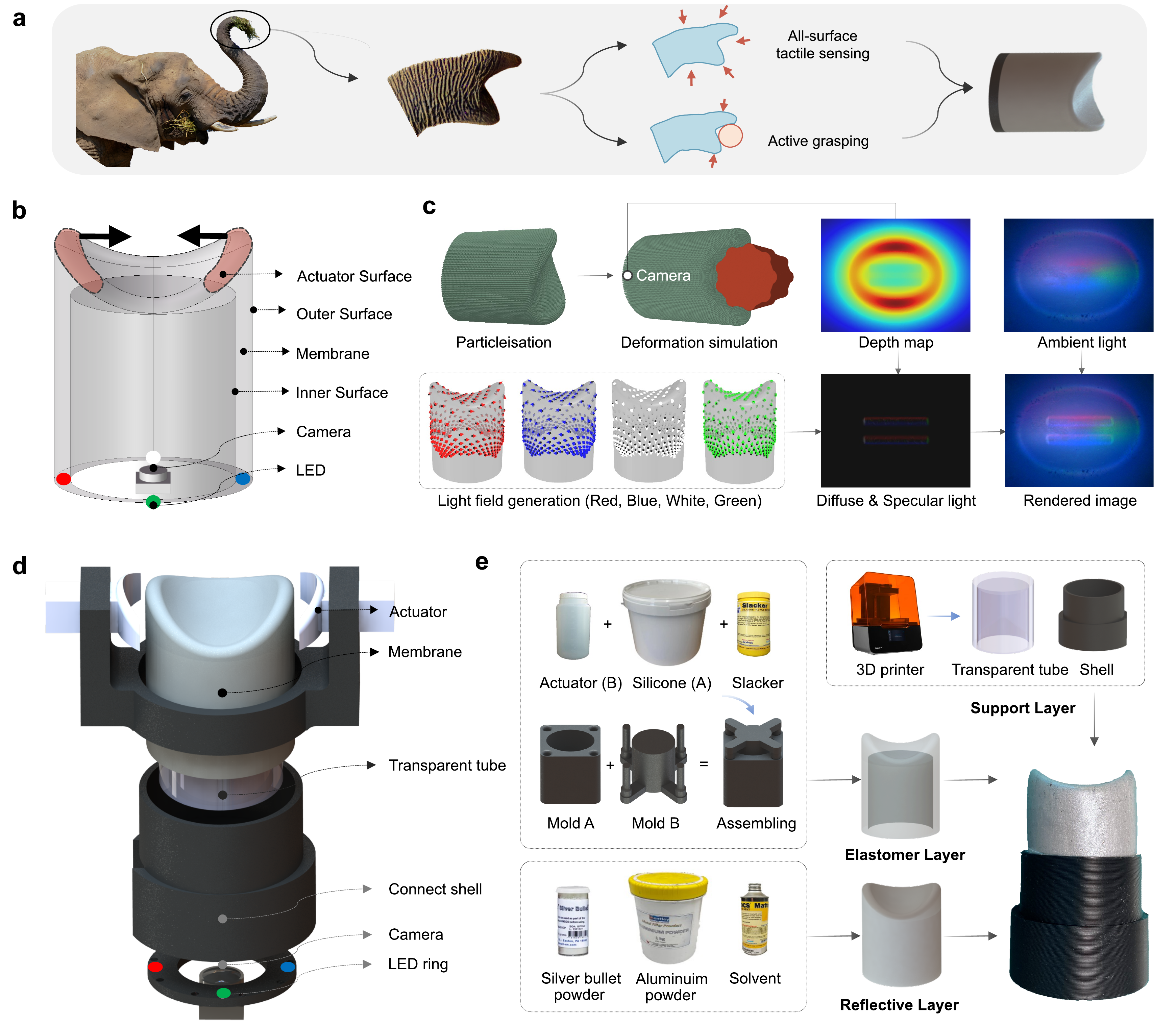}
\caption{Simulation and fabrication of a biomorphic elephant trunk-shaped tactile sensor. \textbf{a}, The inspiration from the elephant trunk, which enables all-surface tactile sensing and active grasping capability. \textbf{b}, Overview of the sensor prototype in simulation. \textbf{c}, Simulation pipeline of the sensor, involving mesh particleization, deformation modelling, and synthetic tactile image generation through light field rendering. \textbf{d}, Exploded view of the real sensor design, showing its modular components, including an actuator, sensor membrane, transparent tube, and integrated optical system. \textbf{e}, Fabrication workflow for the sensor, including mould-based elastomer casting, structural 3D printing, and reflective layer coating.}
\label{fig9}
\end{figure}

As shown in Fig.~\ref{fig9}a, the tip of an elephant’s trunk features two finger-like protrusions that can grasp and manipulate objects like pincers. Its entire surface is covered with skin, providing rich tactile feedback. In contrast, human fingers cannot naturally form a pincer-like structure with a large tactile surface area; instead, they must work together with the palm to achieve similar grasping capabilities. Inspired by the grasping mechanism and advanced tactile sensing of an elephant’s trunk, we designed a biomorphic sensor based on its morphology, where flexible protrusions close inward to execute grasping actions. As illustrated in Fig.~\ref{fig9}b, we first designed a silicone membrane in the shape of an elephant trunk, specified the parameters of the LEDs and cameras, and defined the actuation surface along with the contact conditions between the sensor and the object. We then used SimTac to generate tactile feedback when the sensor interacted with objects, allowing us to evaluate whether the tactile images could accurately and clearly reflect the contact state. Based on the parameter settings in the simulation, we subsequently fabricated a physical prototype of the sensor, as shown in Fig.~\ref{fig9}d, following the manufacturing process illustrated in Fig.~\ref{fig9}e (The detailed parameters of the sensor are provided in Supplementary~\ref{ST3}). 

\begin{figure}[ht!]%
\centering
\includegraphics[width=1\textwidth]{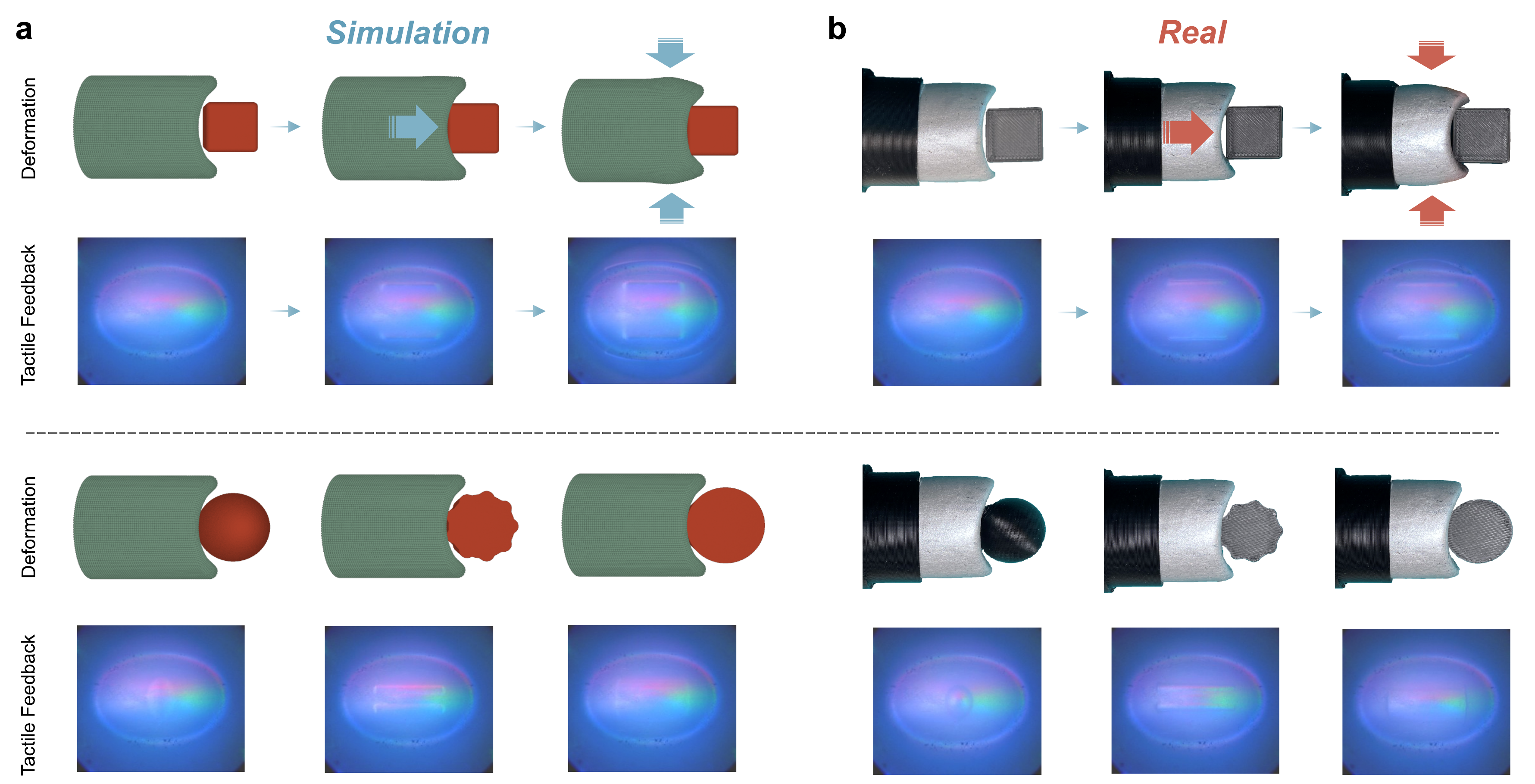}
\caption{Comparison between \textbf{a}, the simulated and \textbf{b}, the real sensor deformation and tactile perception of a biomorphic elephant trunk-shaped tactile sensor.}
\label{fig10}
\end{figure}

Fig.~\ref{fig10} presents the deformation of both the simulated and real elephant trunk when interacting with objects. The trunk first senses the object’s shape through vertical contact, then performs a grasping action by closing its protrusions. The tactile feedback throughout this process is also shown, where the optical response reveals crucial information such as the object's position, shape, and the sensor’s active deformation during grasping. The experimental results demonstrate that SimTac enables the design of innovative biomorphic mechanisms with tactile sensing capabilities by leveraging the interaction mechanisms between biological organisms and their environments. With its capability to simulate tactile responses for diverse biomorphic structures, SimTac holds significant potential for simulating and designing advanced biomorphic sensors and even grippers that integrate both active deformation and high-resolution tactile sensing.

% %\vspace{1em}

\subsubsection{Sim2Real transfer of contact perception.}
\begin{figure}[ht!]%
\centering
\includegraphics[width=1\textwidth]{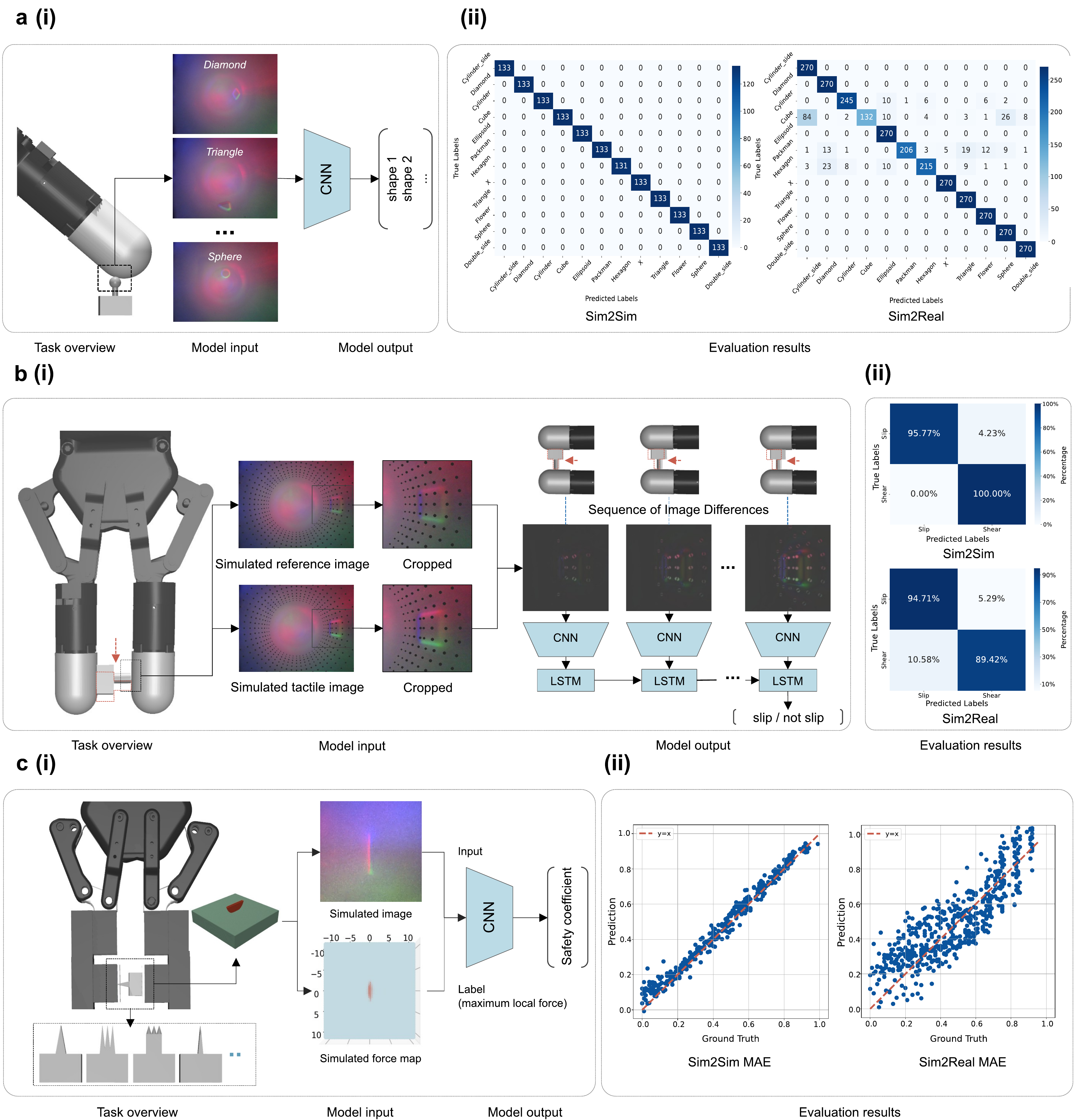}
\caption{Performance evaluation of the tactile-based Sim2Real tasks. \textbf{a}, (\textbf{i}) Pipeline and (\textbf{ii}) results of the object classification Sim2Real task. \textbf{b}, (\textbf{i}) Pipeline and (\textbf{ii}) results of the slip detection Sim2Real task. \textbf{c}, (\textbf{i}) Pipeline and (\textbf{ii}) results of the contact safety assessment Sim2Real task.}
\label{fig11}
\end{figure}

% Setup + principle
The overview of this task is shown in Fig.~\ref{fig11}a(i). We employed a classification model using the ResNet50 architecture~\cite{koonce2021resnet}, which takes tactile images as input and outputs the probability for each shape class (the detailed model structure is provided in Supplementary Fig.~\ref{fig16}a). We established two experimental groups, i.e., Sim2Sim and Sim2Real, to evaluate the model's performance, with the model trained on synthetic data, tested on a synthetic test set in the Sim2Sim group, and assessed on real-world data with zero-shot transfer in the Sim2Real group (detailed in Supplementary~\ref{ST4}).

% Results + Analysis
As shown in Fig.~\ref{fig11}a(ii), the Sim2Sim scenario achieved the highest classification accuracy of 100\%, while the Sim2Real scenario reached an accuracy of 91.3\%. In the Sim2Real scenario, classification failures primarily occurred with indenters of similar shapes. Insufficient contact between the flat indenter and the curved membrane, combined with shape distortion caused by the wide-angle lens, led to the emergence of similar features in the contact area, which prevented the full representation of the indenters' geometric characteristics and resulted in misclassification. Failure cases can be found in Supplementary Fig.~\ref{fig16}b.

%\vspace{1em}

\subsubsection{Sim2Real transfer of slip detection.}
For slip detection tasks, collecting tactile data from real-world scenarios poses significant risks, especially when sensors are exposed to sharp objects or experience prolonged wear. This can damage or even scratch the sensor membrane due to its thin and fragile reflective layer. Therefore, simulating object slip on the sensor surface and training slip detection models with simulated data, while ensuring effective Sim2Real transfer, is essential. The overview of this task is shown in Fig.~\ref{fig11}b(i). We proposed a deep neural network (DNN)-based approach to detect slip, which takes a sequence of eight tactile images as input and outputs a binary classification indicating whether slip has occurred or not (the detailed model structure is provided in Supplementary Fig.~\ref{fig16}b). Two experimental groups were established, i.e., Sim2Sim and Sim2Real, to evaluate the model's performance using synthetic and real data, respectively (detailed in Supplementary~\ref{ST4}).

Supplementary Fig.~\ref{fig17}a presents an example of input image sequences showing both slip and non-slip events in simulated and real-world data. The image sequence demonstrates that the key difference between slip and non-slip lies in the relative motion between the object and the sensor surface. In non-slip cases, the object’s contour and surface markers move synchronously, while in slip cases, the object continues to move as the markers remain nearly stationary, indicating relative motion. The model performance is shown in Fig.~\ref{fig11}b(ii), the prediction accuracy of the Sim2Sim group was 97.89\%, while the Sim2Real group achieved an accuracy of 92.06\%.  

% %\vspace{1em}

\subsubsection{Sim2Real transfer of contact safety assessment.}
Biological organisms rely on tactile perception of local force distribution to assess the safety of physical contact with their environment. For instance, human hands, cat paw pads, and octopus tentacles can detect concentrated pressure when touching sharp objects, triggering rapid withdrawal or other protective responses. This ability is crucial for ensuring safe interactions in robotics as well. However, collecting such tactile data in real-world scenarios carries significant risks. To mitigate these challenges, achieving robust Sim2Real transfer is essential for developing safer and more reliable robotic tactile sensing systems. The overview of this task is shown in Fig.~\ref{fig11}c(i). We constructed a regression model using the ResNet50 architecture~\cite{koonce2021resnet}, which takes tactile images as input and predicts the safety coefficient, a metric used to quantify the magnitude of local contact pressure, ranging from 1 (low local pressure, safe contact) to 0 (high local pressure, high-risk contact). We established two experimental groups to evaluate the performance of the model trained on synthetic data (Sim2Sim) and transferred with zero-shot to real-world scenarios (Sim2Real), detailed in Supplementary~\ref{ST4}.

Fig.~\ref{fig11}c(ii) shows the performance of the trained model in assessing contact safety, achieving an MAE of 0.028 between predicted and ground-truth safety coefficients in Sim2Sim evaluations. In the zero-shot Sim2Real evaluation, the model's MAE was 0.105. When deployed on a real sensor, the simulation-trained model effectively captures the increasing risk of damage as the indentation depth increases, demonstrating its high value for robotic self-protection. The higher Sim2Real error compared to Sim2Sim can be attributed to several factors, as shown in Supplementary Fig.~\ref{fig18}b: Firstly, 3D-printed objects exhibit surface textures absent in simulations, introducing artefacts that affect prediction error. Secondly, as the indentation depth increases, the deformation of the silicone surface alters light distribution, causing the overall image to darken. Thirdly, some real-world data suffer from insufficient indentation, resulting in tactile images that only capture partial contact contours rather than the full contact area. Finally, certain contact poses may cause objects to extend beyond the camera’s field of view, limiting the completeness of tactile information.

\section{Discussion}
% General Description
We introduce SimTac, a physics-based simulator for biomorphic vision-based tactile sensors that can generate synthetic tactile images along with corresponding dense deformation and force fields. The accuracy of SimTac is validated by comparing its outputs with real-world optical and mechanical responses. SimTac demonstrates remarkable flexibility, supporting a wide range of biomorphic geometries, materials, and optical configurations while maintaining high simulation efficiency across varying particle densities and image resolutions. To further demonstrate its capabilities, we design a prototype of a biomorphic elephant trunk sensor through simulation and then fabricate a real sensor based on this design. By comparing the tactile responses, we validate SimTac’s powerful ability to simulate new biomimetic-shaped sensors. This field remains largely unexplored, yet SimTac offers valuable new insights and support for the design and simulation of biomimetic soft robotic tactile systems. Moreover, SimTac shows promising zero-shot sim-to-real transfer performance in representative tactile tasks, including object shape classification, slip detection, and contact safety assessment. These tasks are crucial for enhancing a robot’s ability to perceive object properties, strengthen its interaction with the environment, and protect it from potential harm.

% Necessities
Generally, tactile images can provide rich information about the texture, shape, and pose of the object in contact, which are crucial for identifying its properties. Further deformation and force fields can offer detailed contact information, which is vital for better object interaction. With increasing task complexity, tactile sensors have progressed from simple cubic and hemispherical geometries to finger-like and biomorphic designs. The emergence of such complex morphologies introduces greater challenges for modelling and data generation. This development highlights the need for efficient and accurate simulators capable of producing tactile data across a diverse range of sensor architectures.

% Advantages
For deformation simulation, traditional depth map smoothing methods overlook the physical properties of the membrane, while FEM struggles to generate high-density deformation data in real time. In contrast, combining a particle-based iteration approach with neural networks provides a more efficient approach: the particle-based approach simulates membrane deformation, while neural networks can quickly map this deformation to force fields computed offline by FEM, enabling efficient and accurate generation of dense field data. For optical rendering, data-driven and generative methods typically rely on real sensor data and have yet to be applied to the simulation of complex-shaped sensors. In comparison, physics-based rendering methods, such as path tracing, require more rendering time and are less efficient. The light field-based lighting model rendering method, however, simulates the propagation of light within the sensor membrane of a specific shape, significantly accelerating rendering computations when deployed on a GPU.

% Limitation 
The proposed simulator also has some limitations. The training of the neural network relies on the collection of FEM ground truth data. Although we can fine-tune the predictive model by collecting small batches of FEM data for sensors with the same shape but different materials, data collection for entirely new shapes of sensors still takes a few days at high mesh densities, even though this process is conducted offline and can be accelerated using a GPU.

% Conclusion and impacts.
Overall, our proposed simulator can be applied to tactile sensors with various biomorphic shapes and material parameters. By recording the indices of all particles during particleization, we can control any particle’s motion and achieve active deformation in arbitrary regions, offering potential for simulating actuators. This flexibility eliminates the constraints of fixed sensor geometries, allowing for the integration of biomorphic designs inspired by nature. Therefore, it expands the functionality and application range of tactile sensors, with the potential to drive the development of new robots equipped with tactile sensing.

\section*{Acknowledgments}

\subsection*{Author Contributions} 
X.Z., J.J., and S.L. conceived the method and the experiments and drafted the manuscript. X.Z. designed and constructed the hardware and the simulator, developed fabrication methods, and designed and conducted experiments. X.Z. also collected and analysed the data. Z.C. and Y.Z. helped analyse the data. T.Y. contributed to the model training. D.F.G. contributed to the optical rendering, and J.W. contributed to the revision of the manuscript.

\subsection*{Funding}
This work was supported by the EPSRC project ``ViTac: Visual-Tactile Synergy for Handling Flexible Materials" (EP/T033517/2).

\subsection*{Conflicts of Interest}
The authors declare that they have nocompeting interests.

\subsection*{Data Availability}
Please contact the authors to obtain the data, including the code and dataset.

\printbibliography

@article{li2024biomimetic,
  title={Biomimetic Structure and Surface for Grasping Tasks},
  author={Li, Jingyang and Yin, Fujie and Tian, Yu},
  journal={Biomimetics},
  volume={9},
  number={3},
  pages={144},
  year={2024},
  publisher={MDPI}
}

@article{sun2025soft,
  title={Soft contact simulation and manipulation learning of deformable objects with vision-based tactile sensor},
  author={Sun, Yuhao and Zhang, Shixin and Chen, Zixi and Shen, Zirong and Sun, Fuchun and Stefanini, Cesare and Guo, Di and Luo, Shan and Zhang, Jianwei and Shan, Jianhua and others},
  journal={IEEE Transactions on Automation Science and Engineering},
  year={2025},
  publisher={IEEE}
}

@article{li2022implementing,
  title={Implementing monocular visual-tactile sensors for robust manipulation},
  author={Li, Rui and Peng, Bohao},
  journal={Cyborg and Bionic Systems},
  year={2022},
  publisher={AAAS}
}

@article{fan2025crystaltac,
  title={CrystalTac: Vision-Based Tactile Sensor Family Fabricated via Rapid Monolithic Manufacturing},
  author={Fan, Wen and Li, Haoran and Zhang, Dandan},
  journal={Cyborg and Bionic Systems},
  volume={6},
  pages={0231},
  year={2025},
  publisher={AAAS}
}

@article{lucarotti2013synthetic,
  title={Synthetic and bio-artificial tactile sensing: A review},
  author={Lucarotti, Chiara and Oddo, Calogero Maria and Vitiello, Nicola and Carrozza, Maria Chiara},
  journal={Sensors},
  volume={13},
  number={2},
  pages={1435--1466},
  year={2013},
  publisher={Molecular Diversity Preservation International (MDPI)}
}

@article{gomes2021generation,
  title={Generation of gelsight tactile images for sim2real learning},
  author={Gomes, Daniel Fernandes and Paoletti, Paolo and Luo, Shan},
  journal={IEEE Robotics and Automation Letters},
  volume={6},
  number={2},
  pages={4177--4184},
  year={2021},
  publisher={IEEE}
}

@article{zhao2024fots,
  title={FOTS: A Fast Optical Tactile Simulator for Sim2Real Learning of Tactile-motor Robot Manipulation Skills},
  author={Zhao, Yongqiang and Qian, Kun and Duan, Boyi and Luo, Shan},
  journal={IEEE Robotics and Automation Letters},
  year={2024},
  publisher={IEEE}
}

@article{wang2022tacto,
  title={Tacto: A fast, flexible, and open-source simulator for high-resolution vision-based tactile sensors},
  author={Wang, Shaoxiong and Lambeta, Mike and Chou, Po-Wei and Calandra, Roberto},
  journal={IEEE Robotics and Automation Letters},
  volume={7},
  number={2},
  pages={3930--3937},
  year={2022},
  publisher={IEEE}
}

@article{si2022taxim,
  title={Taxim: An example-based simulation model for gelsight tactile sensors},
  author={Si, Zilin and Yuan, Wenzhen},
  journal={IEEE Robotics and Automation Letters},
  volume={7},
  number={2},
  pages={2361--2368},
  year={2022},
  publisher={IEEE}
}

@article{chen2023tacchi,
  title={Tacchi: A pluggable and low computational cost elastomer deformation simulator for optical tactile sensors},
  author={Chen, Zixi and Zhang, Shixin and Luo, Shan and Sun, Fuchun and Fang, Bin},
  journal={IEEE Robotics and Automation Letters},
  volume={8},
  number={3},
  pages={1239--1246},
  year={2023},
  publisher={IEEE}
}

@article{nguyen2023bioinspiration,
  title={Bioinspiration and biomimetic art in robotic grippers},
  author={Nguyen, Van Pho and Dhyan, Sunil Bohra and Mai, Vu and Han, Boon Siew and Chow, Wai Tuck},
  journal={Micromachines},
  volume={14},
  number={9},
  pages={1772},
  year={2023},
  publisher={MDPI}
}

@inproceedings{sun2023simulation,
  title={Simulation of Vision-based Tactile Sensors with Efficiency-tunable Rendering},
  author={Sun, Yuhao and Zhang, Shixin and Shan, Jianhua and Zhao, Lei and Wang, Xiangbo and Sun, Fuchun and Yang, Yiyong and Fang, Bin},
  booktitle={2023 IEEE International Conference on Robotics and Biomimetics (ROBIO)},
  pages={1--6},
  year={2023}
}

@article{sun2025tacchi,
  title={Tacchi 2.0: A Low Computational Cost and Comprehensive Dynamic Contact Simulator for Vision-based Tactile Sensors},
  author={Sun, Yuhao and Zhang, Shixin and Li, Wenzhuang and Zhao, Jie and Shan, Jianhua and Shen, Zirong and Chen, Zixi and Sun, Fuchun and Guo, Di and Fang, Bin},
  journal={arXiv preprint arXiv:2503.09100},
  year={2025}
}

@ARTICLE{10720429,
  author={Shen, Zirong and Sun, Yuhao and Zhang, Shixin and Chen, Zixi and Sun, Heyi and Sun, Fuchun and Fang, Bin},
  journal={IEEE Robotics and Automation Letters}, 
  title={Simulation of Optical Tactile Sensors Supporting Slip and Rotation Using Path Tracing and IMPM}, 
  year={2024},
  volume={9},
  number={12},
  pages={11218-11225}
}

@article{agarwal2025vision,
  title={Vision-based tactile sensor design using physically based rendering},
  author={Agarwal, Arpit and Wilson, Achu and Man, Timothy and Adelson, Edward and Gkioulekas, Ioannis and Yuan, Wenzhen},
  journal={Communications Engineering},
  volume={4},
  number={1},
  pages={21},
  year={2025},
  publisher={Nature Publishing Group UK London}
}

@inproceedings{xu2023efficient,
  title={Efficient tactile simulation with differentiability for robotic manipulation},
  author={Xu, Jie and Kim, Sangwoon and Chen, Tao and Garcia, Alberto Rodriguez and Agrawal, Pulkit and Matusik, Wojciech and Sueda, Shinjiro},
  booktitle={Conference on Robot Learning},
  pages={1488--1498},
  year={2023}
}

@article{du2024tacipc,
  title={TacIPC: Intersection-and Inversion-free FEM-based Elastomer Simulation For Optical Tactile Sensors},
  author={Du, Wenxin and Xu, Wenqiang and Ren, Jieji and Yu, Zhenjun and Lu, Cewu},
  journal={IEEE Robotics and Automation Letters},
  year={2024},
  publisher={IEEE}
}

@article{si2024difftactile,
  title={DIFFTACTILE: A Physics-based Differentiable Tactile Simulator for Contact-rich Robotic Manipulation},
  author={Si, Zilin and Zhang, Gu and Ben, Qingwei and Romero, Branden and Xian, Zhou and Liu, Chao and Gan, Chuang},
  journal={The International Conference on Learning Representations},
  year={2024}
}

@article{gomes2023beyond,
  title={Beyond flat gelsight sensors: Simulation of optical tactile sensors of complex morphologies for sim2real learning},
  author={Gomes, Daniel Fernandes and Paoletti, Paolo and Luo, Shan},
  journal={Robotics: Science and Systems},
  year={2023}
}

@article{kim2023marker,
  title={Marker-embedded tactile image generation via generative adversarial networks},
  author={Kim, Won Dong and Yang, Sanghoon and Kim, Woojong and Kim, Jeong-Jung and Kim, Chang-Hyun and Kim, Jung},
  journal={IEEE Robotics and Automation Letters},
  volume={8},
  number={8},
  pages={4481--4488},
  year={2023},
  publisher={IEEE}
}

@article{lin2024vision,
  title={Vision-based Tactile Image Generation via Contact Condition-guided Diffusion Model},
  author={Lin, Xi and Xu, Weiliang and Mao, Yixian and Wang, Jing and Lv, Meixuan and Liu, Lu and Luo, Xihui and Li, Xinming},
  journal={arXiv preprint arXiv:2412.01639},
  year={2024}
}

@article{zhong2024tactgen,
  title={TactGen: Tactile Sensory Data Generation via Zero-Shot Sim-to-Real Transfer},
  author={Zhong, Shaohong and Albini, Alessandro and Maiolino, Perla and Posner, Ingmar},
  journal={IEEE Transactions on Robotics},
  year={2024}
}

@INPROCEEDINGS{9561122,
  author={Agarwal, Arpit and Man, Timothy and Yuan, Wenzhen},
  booktitle={2021 IEEE International Conference on Robotics and Automation (ICRA)}, 
  title={Simulation of Vision-based Tactile Sensors using Physics based Rendering}, 
  year={2021},
  volume={},
  number={},
  pages={1-7}}

@article{yuan2017gelsight,
  title={Gelsight: High-resolution robot tactile sensors for estimating geometry and force},
  author={Yuan, Wenzhen and Dong, Siyuan and Adelson, Edward H},
  journal={Sensors},
  volume={17},
  number={12},
  pages={2762},
  year={2017},
  publisher={MDPI}
}

@inproceedings{calli2015ycb,
  title={The ycb object and model set: Towards common benchmarks for manipulation research},
  author={Calli, Berk and Singh, Arjun and Walsman, Aaron and Srinivasa, Siddhartha and Abbeel, Pieter and Dollar, Aaron M},
  booktitle={2015 international conference on advanced robotics (ICAR)},
  pages={510--517},
  year={2015}
}

@inproceedings{donlon2018gelslim,
  title={Gelslim: A high-resolution, compact, robust, and calibrated tactile-sensing finger},
  author={Donlon, Elliott and Dong, Siyuan and Liu, Melody and Li, Jianhua and Adelson, Edward and Rodriguez, Alberto},
  booktitle={2018 IEEE/RSJ International Conference on Intelligent Robots and Systems (IROS)},
  pages={1927--1934},
  year={2018}
}

@inproceedings{padmanabha2020omnitact,
  title={Omnitact: A multi-directional high-resolution touch sensor},
  author={Padmanabha, Akhil and Ebert, Frederik and Tian, Stephen and Calandra, Roberto and Finn, Chelsea and Levine, Sergey},
  booktitle={2020 IEEE International Conference on Robotics and Automation (ICRA)},
  pages={618--624},
  year={2020}
}

@inproceedings{gomes2020geltip,
  title={GelTip: A finger-shaped optical tactile sensor for robotic manipulation},
  author={Gomes, Daniel Fernandes and Lin, Zhonglin and Luo, Shan},
  booktitle={2020 IEEE/RSJ International Conference on Intelligent Robots and Systems (IROS)},
  pages={9903--9909},
  year={2020}
}

@article{ward2018tactip,
  title={The tactip family: Soft optical tactile sensors with 3d-printed biomimetic morphologies},
  author={Ward-Cherrier, Benjamin and Pestell, Nicholas and Cramphorn, Luke and Winstone, Benjamin and Giannaccini, Maria Elena and Rossiter, Jonathan and Lepora, Nathan F},
  journal={Soft robotics},
  volume={5},
  number={2},
  pages={216--227},
  year={2018},
  publisher={Mary Ann Liebert, Inc. 140 Huguenot Street, 3rd Floor New Rochelle, NY 10801 USA}
}

@article{van2020large,
  title={Large-scale vision-based tactile sensing for robot links: Design, modeling, and evaluation},
  author={Van Duong, Lac and others},
  journal={IEEE Transactions on Robotics},
  volume={37},
  number={2},
  pages={390--403},
  year={2020}
}

@article{luu2023simulation,
  title={Simulation, learning, and application of vision-based tactile sensing at large scale},
  author={Luu, Quan Khanh and Nguyen, Nhan Huu and others},
  journal={IEEE Transactions on Robotics},
  volume={39},
  number={3},
  pages={2003--2019},
  year={2023}
}

@inproceedings{church2022tactile,
  title={Tactile sim-to-real policy transfer via real-to-sim image translation},
  author={Church, Alex and Lloyd, John and Lepora, Nathan F and others},
  booktitle={Conference on Robot Learning},
  pages={1645--1654},
  year={2022}
}

@article{lambeta2020digit,
  title={Digit: A novel design for a low-cost compact high-resolution tactile sensor with application to in-hand manipulation},
  author={Lambeta, Mike and Chou, Po-Wei and Tian, Stephen and Yang, Brian and Maloon, Benjamin and Most, Victoria Rose and Stroud, Dave and Santos, Raymond and Byagowi, Ahmad and Kammerer, Gregg and others},
  journal={IEEE Robotics and Automation Letters},
  volume={5},
  number={3},
  pages={3838--3845},
  year={2020},
  publisher={IEEE}
}

@INPROCEEDINGS{10256475,
  author={Cui, Shaowei and Zhang, Chaofan and Shao, Zhihao and Dong, Chengju and Wang, Yu and Wang, Shuo and Tian, Yuan},
  booktitle={2023 IEEE 13th International Conference on CYBER Technology in Automation, Control, and Intelligent Systems (CYBER)}, 
  title={Contact Volumetric Mesh Simulation of GelStereo Visuotactile Sensors}, 
  year={2023},
  volume={},
  number={},
  pages={1363-1367}}

@inproceedings{cui2023tactile,
  title={Tactile imprint simulation of gelstereo visuotactile sensors},
  author={Cui, Shaowei and Wang, Yu and Wang, Shuo and Li, Qian and Wang, Rui and Zhang, Chaofan},
  booktitle={2023 IEEE International Conference on Mechatronics and Automation (ICMA)},
  pages={650--656},
  year={2023}
}

@article{zhang2024tacpalm,
  title={Tacpalm: A soft gripper with a biomimetic optical tactile palm for stable precise grasping},
  author={Zhang, Xuyang and Yang, Tianqi and Zhang, Dandan and Lepora, Nathan F},
  journal={IEEE Sensors Journal},
  year={2024}
}

@article{sun2022soft,
  title={A soft thumb-sized vision-based sensor with accurate all-round force perception},
  author={Sun, Huanbo and Kuchenbecker, Katherine J and Martius, Georg},
  journal={Nature Machine Intelligence},
  volume={4},
  number={2},
  pages={135--145},
  year={2022},
  publisher={Nature Publishing Group UK London}
}

@article{zhang2024rotip,
  title={Rotip: A finger-shaped tactile sensor with active rotation},
  author={Zhang, Xuyang and Jiang, Jiaqi and Luo, Shan},
  journal={arXiv preprint arXiv:2410.01085},
  year={2024}
}

@article{jiang2024rotipbot,
  title={RoTipBot: Robotic Handling of Thin and Flexible Objects using Rotatable Tactile Sensors},
  author={Jiang, Jiaqi and Zhang, Xuyang and Gomes, Daniel Fernandes and Do, Thanh-Toan and Luo, Shan},
  journal={IEEE Transactions on Robotics},
  year={2025}
}

@article{lepora2022digitac,
  title={Digitac: A digit-tactip hybrid tactile sensor for comparing low-cost high-resolution robot touch},
  author={Lepora, Nathan F and Lin, Yijiong and Money-Coomes, Ben and Lloyd, John},
  journal={IEEE Robotics and Automation Letters},
  volume={7},
  number={4},
  pages={9382--9388},
  year={2022},
  publisher={IEEE}
}

@article{chen2019ultrahigh,
  title={An ultrahigh resolution pressure sensor based on percolative metal nanoparticle arrays},
  author={Chen, Minrui and Luo, Weifeng and Xu, Zhongqi and Zhang, Xueping and Xie, Bo and Wang, Guanghou and Han, Min},
  journal={Nature communications},
  volume={10},
  number={1},
  pages={4024},
  year={2019},
  publisher={Nature Publishing Group UK London}
}

@article{boutry2018hierarchically,
  title={A hierarchically patterned, bioinspired e-skin able to detect the direction of applied pressure for robotics},
  author={Boutry, Clementine M and Negre, Marc and Jorda, Mikael and Vardoulis, Orestis and Chortos, Alex and Khatib, Oussama and Bao, Zhenan},
  journal={Science Robotics},
  volume={3},
  number={24},
  pages={eaau6914},
  year={2018},
  publisher={American Association for the Advancement of Science}
}

@article{bai2020stretchable,
  title={Stretchable distributed fiber-optic sensors},
  author={Bai, Hedan and Li, Shuo and Barreiros, Jose and Tu, Yaqi and Pollock, Clifford R and Shepherd, Robert F},
  journal={Science},
  volume={370},
  number={6518},
  pages={848--852},
  year={2020},
  publisher={American Association for the Advancement of Science}
}

@article{akinola2024tacsl,
  title={Tacsl: A library for visuotactile sensor simulation and learning},
  author={Akinola, Iretiayo and Xu, Jie and Carius, Jan and Fox, Dieter and Narang, Yashraj},
  journal={IEEE Transactions on Robotics},
  year={2025}
}

@article{zhou2018open3d,
  title={Open3D: A modern library for 3D data processing},
  author={Zhou, Qian-Yi and Park, Jaesik and Koltun, Vladlen},
  journal={arXiv preprint arXiv:1801.09847},
  year={2018}
}

@incollection{phong1998illumination,
  title={Illumination for computer generated pictures},
  author={Phong, Bui Tuong},
  booktitle={Seminal graphics: pioneering efforts that shaped the field},
  pages={95--101},
  year={1998}
}

@inproceedings{liu2015sparse,
  title={Sparse convolutional neural networks},
  author={Liu, Baoyuan and Wang, Min and Foroosh, Hassan and Tappen, Marshall and Pensky, Marianna},
  booktitle={Proceedings of the IEEE conference on computer vision and pattern recognition},
  pages={806--814},
  year={2015}
}

@inproceedings{hore2010image,
  title={Image quality metrics: PSNR vs. SSIM},
  author={Hore, Alain and Ziou, Djemel},
  booktitle={2010 20th international conference on pattern recognition},
  pages={2366--2369},
  year={2010}
}

@article{koonce2021resnet,
  title={ResNet 50},
  author={Koonce, Brett and Koonce, Brett},
  journal={Convolutional neural networks with swift for tensorflow: image recognition and dataset categorization},
  pages={63--72},
  year={2021},
  publisher={Springer}
}

@article{tammina2019transfer,
  title={Transfer learning using vgg-16 with deep convolutional neural network for classifying images},
  author={Tammina, Srikanth},
  journal={International Journal of Scientific and Research Publications (IJSRP)},
  volume={9},
  number={10},
  pages={143--150},
  year={2019}
}

@inproceedings{deng2009imagenet,
  title={Imagenet: A large-scale hierarchical image database},
  author={Deng, Jia and Dong, Wei and Socher, Richard and Li, Li-Jia and Li, Kai and Fei-Fei, Li},
  booktitle={2009 IEEE conference on computer vision and pattern recognition},
  pages={248--255},
  year={2009}
}

@article{graves2012long,
  title={Long short-term memory},
  author={Graves, Alex and Graves, Alex},
  journal={Supervised sequence labelling with recurrent neural networks},
  pages={37--45},
  year={2012},
  publisher={Springer}
}

@article{he2023tacmms,
  title={TacMMs: Tactile mobile manipulators for warehouse automation},
  author={He, Zhuochao and Zhang, Xuyang and Jones, Simon and Hauert, Sabine and Zhang, Dandan and Lepora, Nathan F},
  journal={IEEE Robotics and Automation Letters},
  volume={8},
  number={8},
  pages={4729--4736},
  year={2023},
  publisher={IEEE}
}

@inproceedings{choy20194d,
  title={4d spatio-temporal convnets: Minkowski convolutional neural networks},
  author={Choy, Christopher and Gwak, JunYoung and Savarese, Silvio},
  booktitle={Proceedings of the IEEE/CVF conference on computer vision and pattern recognition},
  pages={3075--3084},
  year={2019}
}

@inproceedings{luo2018vitac,
  title={Vitac: Feature sharing between vision and tactile sensing for cloth texture recognition},
  author={Luo, Shan and Yuan, Wenzhen and Adelson, Edward and Cohn, Anthony G and Fuentes, Raul},
  booktitle={2018 IEEE International Conference on Robotics and Automation (ICRA)},
  pages={2722--2727},
  year={2018}
}

@incollection{jiang2022robotic,
  title={Robotic perception of object properties using tactile sensing},
  author={Jiang, Jiaqi and Luo, Shan},
  booktitle={Tactile Sensing, Skill Learning, and Robotic Dexterous Manipulation},
  pages={23--44},
  year={2022},
  publisher={}
}

@inproceedings{dong2017improved,
  title={Improved gelsight tactile sensor for measuring geometry and slip},
  author={Dong, Siyuan and Yuan, Wenzhen and Adelson, Edward H},
  booktitle={IEEE/RSJ International Conference on Intelligent Robots and Systems (IROS)},
  pages={137--144},
  year={2017}
}

@article{she2021cable,
  title={Cable manipulation with a tactile-reactive gripper},
  author={She, Yu and Wang, Shaoxiong and Dong, Siyuan and Sunil, Neha and Rodriguez, Alberto and Adelson, Edward},
  journal={The International Journal of Robotics Research},
  volume={40},
  number={12-14},
  pages={1385--1401},
  year={2021},
  publisher={SAGE Publications Sage UK: London, England}
}

@article{bern2000mesh,
  title={Mesh generation},
  author={Bern, Marshall W and Plassmann, Paul E},
  journal={Handbook of computational geometry},
  volume={38},
  year={2000}
}

@article{decherchi2013general,
  title={A general and robust ray-casting-based algorithm for triangulating surfaces at the nanoscale},
  author={Decherchi, Sergio and Rocchia, Walter},
  journal={PloS one},
  volume={8},
  number={4},
  pages={e59744},
  year={2013},
  publisher={Public Library of Science San Francisco, USA}
}

@inproceedings{calandra2017feeling,
  title={The Feeling of Success: Does Touch Sensing Help Predict Grasp Outcomes?},
  author={Calandra, Roberto and Owens, Andrew and Upadhyaya, Manu and Yuan, Wenzhen and Lin, Justin and Adelson, Edward H and Levine, Sergey},
  booktitle={Conference on Robot Learning},
  pages={314--323},
  year={2017}
}

@article{chen2024transforce,
  title={TransForce: Transferable Force Prediction for Vision-based Tactile Sensors with Sequential Image Translation},
  author={Chen, Zhuo and Ou, Ni and Zhang, Xuyang and Luo, Shan},
  journal={2025 IEEE International Conference on Robotics and Automation (ICRA)},
  year={2025}
}

@inproceedings{billen2016line,
  title={Line sampling for direct illumination},
  author={Billen, Niels and Dutr{\'e}, Philip},
  booktitle={Computer Graphics Forum},
  volume={35},
  number={4},
  pages={45--55},
  year={2016}
}

@article{de2020material,
  title={Material point method after 25 years: Theory, implementation, and applications},
  author={De Vaucorbeil, Alban and Nguyen, Vinh Phu and Sinaie, Sina and Wu, Jian Ying},
  journal={Advances in applied mechanics},
  volume={53},
  pages={185--398},
  year={2020},
  publisher={Elsevier}
}

\newpage

\section*{Supplementary Materials}

% \noindent \textbf{This Supplementary Materials file includes:}

% \noindent Supplementary Text 1.  Data Collection Setup

% \noindent Supplementary Text 2. MPM Iteration Method

% \noindent Supplementary Text 3.  Sim2Real Task Setup

% \noindent Supplementary Text 4. FEM Model Setup for Generating Ground-Truth Data

% \noindent Supplementary Figure 12 to 18.

% \noindent Supplementary Video 1 to 5.

% \setcounter{subsection}{0}

\subsection{MPM iteration method}\label{ST2} 
We employ the Material Point Method (MPM)~\cite{chen2023tacchi}, a particle-based simulation approach, to model the deformation of sensor skin upon contact with an object. MPM discretises objects into particles that carry properties, such as mass and velocity, and updates their state through an iterative process of exchanging properties between particles and virtual grids. This process enables a realistic simulation of deformation over time. The grids remain fixed, while properties are transferred between particles and grid nodes to facilitate the simulation. Each time step consists of four stages: (1) Particle-to-Grid (P2G): The particle properties are interpolated onto neighbouring grid nodes; (2) Grid Operations: momentum transferred from nearby particles are computed, and the velocities of the grid nodes are updated accordingly; (3) Grid-to-Particle (G2P): the updated velocities are interpolated back to the particles; and (4) Particle Operations: particle positions are updated, with boundary conditions applied. 

% %\vspace{1em}

\subsubsection{Initialisation.} 
The sensor skin and object are discretised into \( m \) particles, with a virtual grid consisting of \( n \) fixed nodes to transfer the particle properties. For the \( p \)-th particle, its position is indicated as \( {x}_{p} \in \mathbb{{R}}^3 \), and its velocity is given by \( {v}_{p} \in \mathbb{{R}}^3 \). Affine velocity \( C_p \in \mathbb{{R}}^{3\times3}\) is introduced to capture the non-linear local deformations within the object. A deformation field \( {\varphi}_p: \mathbb{{R}}^3 \to \mathbb{{R}}^3 \) is used to record the initial and final positions of each particle in each time step. The deformation gradient \( {F}_p \in \mathbb{{R}}^{3\times3} \) represents the extent of deformation relative to its initial state, which can be derived as:
\begin{equation}
F_p = \frac{\partial \varphi_p}{\partial x_p} (x_p)
\end{equation}

% %\vspace{1em}

\subsubsection{Particle-to-Grid.} This process maps particle properties to grid nodes through interpolation, using quadratic B-spline weighting to compute the influence of each particle on the surrounding grid nodes. The mass and the momentum of the particles are then accumulated onto the corresponding grid nodes based on these weights. The mass \( M_i \) of the \(i\)-th grid node is:
\begin{equation}
M_i = \sum_{j \in \mathbb{G}_i} \sum_{p \in \mathbb{P}_j} w_{jp} m_p
\end{equation}
where \( \mathbb{G}_i \) denote the \(3\times3\times3\) grid nodes containing the \( i \)-th grid node and its neighboring grid nodes, \( \mathbb{P}_j \) denote the particles inside the \( j \)-th grid, and \( w_{ij} \) denote the weight parameter between the \( i \)-th grid node and the \( j \)-th particle, calculated using quadratic B-Spline interpolation, and \( m_p \) denote the mass of the \( p \)-th particle. 

The momentum \( {MG}_i \) of the \( i \)-th grid node can be obtained by calculating the momentum resulted from the particle motion \( {MM}_i \) and the momentum resulted from the elasticity
\( {ME}_i \):
\begin{equation}
MG_i = MM_i + ME_i
\end{equation}

\( MM_i \) is calculated by collecting the velocity and affine velocity of the nearby particles (A temporal superscript is utilised to distinguish the parameters at each different time step, e.g., \(v_p^{(k)}\) denoting the velocity at time step \( k\)): 
\begin{equation}
MM_i = \sum_{j \in \mathbb{G}_i} \sum_{p \in \mathbb{P}_j} w_{jp} \left( m_p v_p^{(k)} + C_p^{(k)} (X_j - x_p^{(k)}) \right)
\end{equation}
where \( {X}_j \) denotes the position of the \( j \)-th neighboring node of the \( i \)-th grid node.

\( ME_i \) is obtained by accumulating the elastic stress from neighbouring particles:
\begin{equation}
ME_i = -\Delta t \sum_{j \in \mathbb{G}_i} \sum_{p \in \mathbb{P}_j} \frac{4}{\Delta X^2} w_{jp} V_p^0 S_p^{(k)} (X_j - x_p^{(k)})
\end{equation}
where \( \Delta t \) is the time interval between two adjacent steps,  \( \Delta X \) is the grid node interval,  \( V_p^0 \) denotes the initial particle volume,  \( S_p \) is the elasticity force for the \( p \)-th particle. 

%\vspace{1em}

\subsubsection{Grid Operation.} The velocity on the \(i\)-th grid node can be obtained given the grid node momentum and the grid node mass:
\begin{equation}
V_i = \frac{MG_i}{M_i}
\end{equation}
Note that the grid node velocity is only for later parameter updates, and the position of the grid will not change in the simulation.

%\vspace{1em}

\subsubsection{Grid-to-Particle.} The states of the particles are then updated with the previous states of the grid nodes and the particles. The velocity  \(v_p^{(k+1)}\), the affine velocity \(C_p^{(k+1)}\) and deformation gradient \(F_p^{(k+1)}\) at time step \(k+1\) can be obtained by:
\begin{equation}
v_p^{(k+1)} = \sum_{i \in \mathbb{G}_p^\prime} w_{ip} V_i^{(k)}
\end{equation}

\begin{equation}
C_p^{(k+1)} = \frac{\Delta X^2}{\Delta t} \sum_{i \in \mathbb{G}_p^\prime} w_{ip} v_p^{(k+1)} \frac{X_i - x_p^{(k)}}{\Delta X}
\end{equation}

\begin{equation}
F_p^{(k+1)} = \left( I + \Delta t C_p^{(k+1)} \right) F_p^{(k)}
\end{equation}
where \( G_i^\prime \) represents the \(3\times3\times3\) grid nodes surrounding the \( p \)-th particle.

%\vspace{1em}

\subsubsection{Particle Operation.} We will apply specific boundary conditions to constrain the motion of certain particles. Particles within the rigid body \(\mathbb{R}\) will share the same velocity \( v \) (e.g., the rigid indenter and sensor actuator), while those in the fixed region \(\mathbb{B}\) will have a velocity of \(0\) (e.g., the inner support layer of the sensor membrane if the sensor is fixed):
\begin{equation}
v_{p}^{(k+1)} =
\begin{cases}
v & \text{if } p \in \mathbb{I}, \\
0 & \text{if } p \in \mathbb{B},
\end{cases}
\end{equation}

Finally, the position of each particle \(x_{p}^{(k+1)}\) at \(k+1\) time step can be calculated by:
\begin{equation}
x_{p}^{(k+1)} = x_{p}^{(k)} + \Delta t \, v_{p}^{(k+1)}
\end{equation}

The entire flow chart of the MPM iteration is sketched in Algorithm~\ref{alg3}.  

\begin{algorithm}
\caption{Material Point Method Iteration}
\label{alg3}
\begin{algorithmic}[1]
\State \textbf{Input:} Particles of the sensor skin and the objects (\( P \)); Number of the grid nodes (\( n\)); Material parameters of the skin (\( E \) and \( \nu \)); Relative velocity of the contact \( v_r \); Index of interested particles (boundary conditions, contact surfaces, actuator surfaces).
\State \textbf{Output:} Positions of the deformed particles.
\State Discrete the model into particles using the free mesh algorithm.
\State Initialise the values of \( x^{(0)}_p \), \( F^{(0)}_p \), \( C^{(0)}_p \), and \( v^{(0)}_p \).
\State Divide the simulation space into grids.
\While{not terminal}
    \For{each grid \( i \)}
        \State Collect the mass \(M_i\) and momentum \(MG_i\) of grid \( i \) by Eq. (10-13).
        \State Update the grid velocity \(V_i\) by Eq. (14).
    \EndFor
    \For{each particle \( p \)}
        \State Update parameters \( v^{(k+1)}_p \), \( F^{(k+1)}_p \), and \( C^{(k+1)}_p \) by Eq. (15-17). \Comment{\( k\) represents the \( k\)-th time step}
    \EndFor
    \State Boundary conditions applied by Eq. (18).
    \State Particle position updated by Eq. (19).
    \State Terminal check: whether the target pose of the indenter/sensor is achieved.
\EndWhile
\end{algorithmic}
\end{algorithm}

\subsection{Data collection setup}\label{ST1}

We mainly collected tactile data from the finger-shaped GelTip sensor~\cite{gomes2020geltip} and the cubic-shaped GelSight sensor~\cite{yuan2017gelsight}. The GelTip dataset was used to evaluate the accuracy of the simulator and validate Sim2Real tasks such as contact object classification and slip detection, while the GelSight dataset was solely utilised for contact safety assessment. 

% %\vspace{1em}
 
\subsubsection{Data collection on GelTip sensor.}
We set up a data collection platform as shown in Fig.~\ref{fig4}a. The GelTip sensor is securely mounted on the table, while the indenter and the ATI NANO 17 force sensor are attached to the UR5e robot arm for different contact positions and orientations. The sensor has multiple slots in its body and can be rotated in 90-degree increments for flexible positioning by selecting different slots to connect to the table base. We also replicate the same sensor and indenter configuration in both the proposed simulator and the FEM, collecting data in both environments for comparison. A total of 14 distinct indenter shapes were utilised, with 10 shapes in the seen group for the STN model training and 4 in the unseen group for testing.

The data acquisition process is shown in Fig.~\ref{fig4}d. The contact path extends from the tip to the base of the membrane, with contact points distributed at 90-degree intervals around the circumference to ensure full coverage of the sensor surface. In each acquisition cycle, the robotic arm first positions the indenter at each contact point, maintaining a perpendicular orientation to the contact surface, and then performs a sequence of actions: normal contact, shear motion, and normal release. During both the normal contact and shear motion phases, tactile images and force data are recorded synchronously at a frequency of 30 Hz.  The pseudocode for the data acquisition process is provided in Algorithm \ref{alg1}.

The data collection parameters are defined as follows and are illustrated in Fig.~\ref{fig4}d: the indenter rotates within the tip region from \(0^\circ\) to \(90^\circ\) with an interval of \(\Delta \alpha = 10^\circ\), and moves within the base region from \(0\) mm to \(12\) mm with a step size of \(\Delta d = 1\) mm. Four contact paths are evenly distributed around the sensor circumference at \(90^\circ\) intervals. At each contact position, the indentation depth \(z\) ranges from \(0\) mm to \(1\) mm with a depth increment of \(\Delta z = 0.1\) mm. For each indentation depth, shear motion is applied in \(8\) different directions, with an angular interval of \(\Delta \beta = 45^\circ\). The shear displacement along the \(x\) or \(y\) axis varies from \(0\) mm to \(0.5\) mm, with a step size of \(\Delta x = 0.1\) mm and \(\Delta y = 0.1\) mm. The data collection setup is shared across the real world, SimTac, and FEM simulation environments.
%a total of 9,960 pairs of tactile images and force data were collected across all environments.

\begin{algorithm}
\caption{Data Collection on GelTip Sensor}
\label{alg1}
\begin{algorithmic}[1]
\raggedright
\State \textbf{Input:} Surface contact points $P = \{P_1, ..., P_m\}$, where $P_i = \{p_1, ..., p_n\}$ contains all contact points in depth, $p_j = \{x, y, z\}$ denotes the coordinate of a contact point, 8 shear directions in each $p_j$, $S =  \{s_1, ..., s_8\}$ 
\Procedure{MoveToContactPoints}{$P, m$}
    \For{$i = 0 \dots m$}
        \State $p_0 \gets P_i[0]$ \Comment{origin point $p_0$ without contact}
        \State \textit{moveTo}($p_0$)
        \State $n_{p0} \gets \frac{p_0}{\|p_0\|}$ 
        \Comment{calculate normal vector of contact point $p_0$}
        \State \textit{adjustOrientation}($n_{p0}$)
        \Comment{make the indenter perpendicular to the sensor surface}
        \For{$j = 1 \dots n$}
            \State $p_1 \gets P_{ij}$
            \State \textit{moveTo}($p_1$), \textit{$record\_image\_and\_force$}() \Comment{Normal increases}
            \For{$k = 1 \dots 8$}
                \State \textit{moveTo}($s_k$), \textit{$record\_image\_and\_force$}() \Comment{Shear motion}
                \State \textit{moveTo}($p_0$) \Comment{Normal decreases}
            \EndFor
        \EndFor
    \EndFor
\EndProcedure
\end{algorithmic}
\end{algorithm}

% %\vspace{1em}

\subsubsection{Data collection on GelSight sensor.}
\begin{figure}[ht!]%
\centering
\includegraphics[width=1\textwidth]{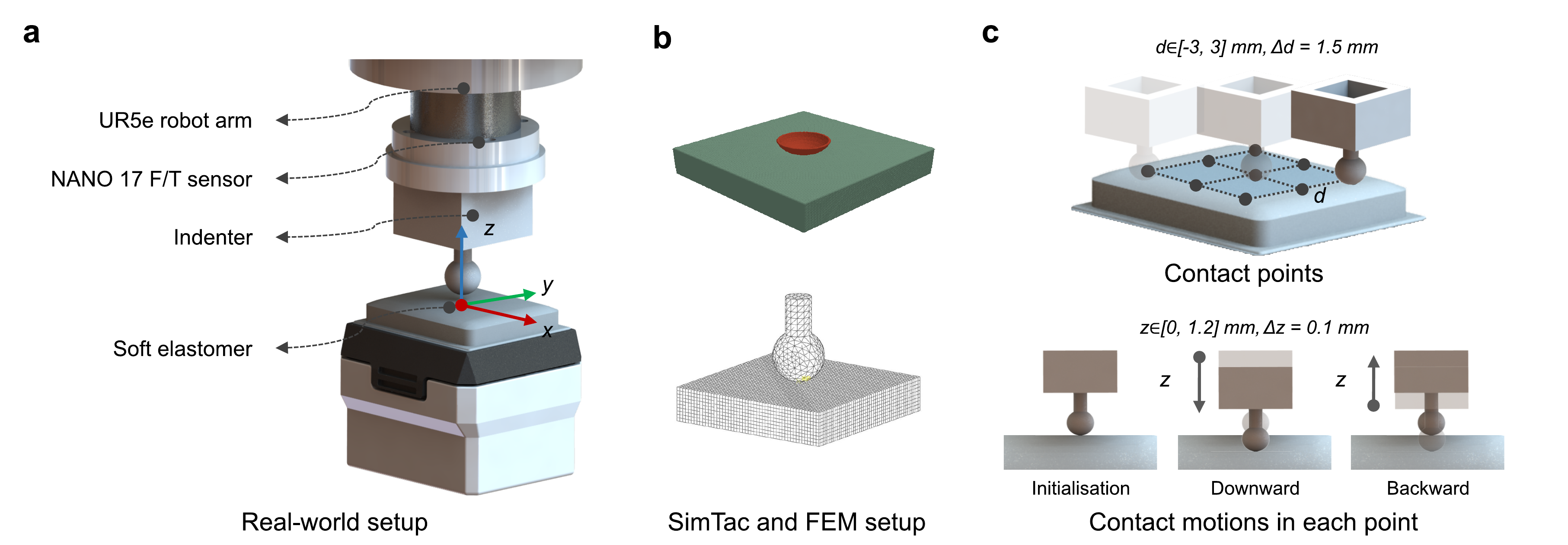}
\caption{Data collection setup on the GelSight sensor. \textbf{a-b}, Real-world, SimTac and FEM data collection platform for the GelSight sensor. \textbf{c}, The movement trajectory of the indenter during the GelSight data collection process.}
\label{Gelsight}
\end{figure}

The data collection platform for the GelSight sensor, shown in Fig.~\ref{Gelsight}a, closely resembles the setup used for the GelTip sensor. We used the same indenters as the GelTip sensor and replicated the setup in both the proposed simulator and the FEM environment, as shown in Fig.~\ref{Gelsight}b. The data collection process is illustrated in Fig.~\ref{Gelsight}c and in Algorithm \ref{alg2}, where the contact points are distributed in a square grid pattern, centred around the sensor surface, ranging from \(-3\) mm to \(3\) mm, with a spacing of \(\Delta d = 1.5\) mm. The contact depth \(z\) varies from \(0\) to \(1.2\) mm, with a depth interval of \(\Delta z = 0.1\) mm. The indenter applies only normal contact, without any shear motion. All other settings remain consistent with those used in the data collection for the GelTip sensor.

\begin{algorithm}
\caption{Data Collection on the GeSight Sensor}
\label{alg2}
\begin{algorithmic}[2]
\raggedright
\State \textbf{Input:} Surface contact points $P = \{P_1, ..., P_m\}$, where $P_i = \{p_1, ..., p_n\}$ contains all contact points in depth, $p_i = \{x, y, z\}$ denotes the coordinate of a contact point.
\Procedure{MoveToContactPoints}{$P, m$}
    \For{$i = 0 \dots m$}
        \State $p_0 \gets P_i[0]$ \Comment{origin point $p_0$ without contact}
        \State \textit{moveTo}($p_0$)
        \For{$j = 1 \dots n$}
            \State $p_1 \gets (p_0[0], p_0[1], P_{ij}[2])$
            \State \textit{moveTo}($p_1$), \textit{$record\_image\_and\_force$}() \Comment{Normal increases}
            \State \textit{moveTo}($p_0$), \textit{$record\_image\_and\_force$}() \Comment{Normal decreases}
        \EndFor
    \EndFor
\EndProcedure
\end{algorithmic}
\end{algorithm}

\subsection{The elephant-trunk-shaped sensor}\label{ST3}
\subsubsection{Simulated sensor setup.} 
We first designed and implemented a sensor prototype within the simulation environment, as illustrated in Fig.~\ref{fig9}b. The surfaces of the two protrusions at the sensor tip were designated as actuator surfaces, where specific particle velocities were applied to achieve opening and closing motions. The outer surface of the elastomer was assigned as the reflective layer, whose deformations were leveraged for optical rendering to generate tactile images. The inner surface was defined as the support layer, with particle velocities constrained to zero. A 120-degree wide-angle camera was positioned at the base of the membrane to capture deformations occurring within the tip region. A ring of red, white, blue, and green LEDs was arranged around the membrane base to provide structured internal illumination.

The simulation pipeline is outlined in Fig.~\ref{fig9}c. We first employed the Unstructured Mesh Algorithm~\cite{bern2000mesh} to generate a $C3D4$ mesh (four-node tetrahedral elements) with a particle spacing of \( h = 0.4 \)~mm to discretise the sensor membrane. The nodal coordinates of the generated mesh were extracted to initialise the positions of the membrane particles. The object in contact with the sensor was modelled as a rigid body and discretised via uniform sampling using \textit{Open3D}~\cite{zhou2018open3d} into \( 1 \times 10^6 \) particles, with all particles sharing the same velocity. The material properties of the membrane were defined by a Young’s modulus of \( E = 1.45 \times 10^5 \)~kPa and a Poisson’s ratio of \( v = 0.45 \). The simulation environment was configured with 256 grid nodes over a total grid length of 70 mm. After initialisation, linear and non-linear light fields were generated offline according to the geometric model of the sensor membrane. Optical rendering was subsequently performed using the depth maps obtained from deformation simulations. The ambient background images and the foreground images generated by diffuse and specular reflections were composited to synthesise the final tactile images.

%\vspace{1em}

\subsubsection{Real sensor setup.} The real sensor consists of four key components: the actuator module, the optical module, the skin module and the lighting module, as shown in Fig.~\ref{fig9}d. The overall dimensions of the sensor are 34 mm $\times$ 26 mm $\times$ 50 mm. The actuator module comprises two sliders, which are driven by an external force to enable opening and closing movements. The optical module includes a 5MP-OV5640-USB camera, capable of capturing images at a 1920 $\times$ 1080 resolution with a 30 Hz frame rate. The camera is equipped with a 120-degree distortion-free fixed-focus lens with a 1.5 mm focal length. The skin module consists of three layers: a transparent rigid support layer, a transparent elastomer layer, and a reflective coating layer. 

The support layer provides structural reinforcement and limits deformation, the elastomer layer enables deformation while facilitating light propagation, and the reflective coating blocks external light while enhancing internal reflection. The rigid support layer is fabricated from a transparent resin using a Formlabs Form-3 3D printer and polished to improve transparency. The elastomer layer is cast using a 3D-printed mould, with the mould surface polished for better clarity. The elastomer is made of silicone (XP-565 from Silicones, Inc.), mixed at a 1:15 ratio. The reflective coating layer consists of a membrane mixed with aluminium powder and reflective pigment. The lighting module consists of an LED ring, which serves as the internal light source. We designed a programmable ring-shaped LED system (WS2812 control chip, 1615 SMD LEDs) that allows precise control over the light sources through adjustable RGB values and intensity settings, ensuring accurate calibration with the simulated lighting conditions. The LEDs are arranged in a red-white-blue-green sequence, with equal brightness levels of 1:1:1:1. The fabrication process is detailed in Fig.~\ref{fig9}e.

\subsection{Sim2Real task setup}\label{ST4}

\subsubsection{Contact object classification setup.}
We used the ResNet50~\cite{koonce2021resnet} architecture as our machine learning model, which uses tactile images as input and outputs the probability for each shape class. Following the data collection method described in Supplementary~\ref{ST1}, only vertical contact was applied, without any shear motion. A total of 8,016 tactile image pairs were collected from the contact of different-shaped indenters with the GelTip sensor in both real and simulated environments. These image pairs were divided into training, validation, and test sets in a ratio of 3:1:1. During the training process, all parameters of the convolutional layer are initialised using a pre-trained model on ImageNet~\cite{deng2009imagenet} to enhance feature extraction capabilities. The classification model was trained with a batch size of 32 for 30 epochs using the Adam optimiser with a learning rate of 0.001 to minimise the mean squared error (L1 loss). 
% Augment

%\vspace{1em}

\subsubsection{Slip detection setup.}
We propose a deep neural network (DNN)-based approach for slip detection, which takes a sequence of eight tactile images as input and outputs a binary classification indicating whether slip has occurred or not. Following the data collection method described in Supplementary~\ref{ST1}, we collected a total of 832 tactile sequences involving shear contact between various indenter shapes and the GelTip sensor, in both simulated and real-world settings. The contact area was restricted to the sidewall region, which is the primary contact area when the sensor grasps an object. The dataset is balanced, with an equal number of slip and non-slip samples (1:1 ratio). To accurately label the data, we first replicated the real-world experimental setup and shear loading process in FEM, enabling the recording of tangential forces applied to the sensor. The onset of slip is defined as the point where the tangential force curve transitions from increasing to decreasing. The displacement at this turning point is used as the slip detection threshold: sequences with displacements below this threshold are labelled as non-slip, while those above are labelled as slip. This thresholding method is applied to all simulated sequences. For the real-world data, we mounted a force sensor on the indenter to directly measure tangential force, using the same criterion for consistent labelling.

The model is trained exclusively on simulated data, with 60\% used for training, 20\% for validation, and 20\% for testing. All real-world data is reserved for testing to assess sim-to-real transfer performance. The network architecture consists of a convolutional neural network (CNN) for spatial feature extraction and a recurrent neural network (RNN) for temporal decision-making. Specifically, we use a VGG-19 backbone~\cite{tammina2019transfer} pre-trained on ImageNet~\cite{deng2009imagenet} to encode tactile features, followed by a long short-term memory (LSTM) network~\cite{graves2012long} to model temporal dependencies across frames. The model is trained using the Adam optimiser with a learning rate of 0.0005, a batch size of 4, and cross-entropy loss, for 15 epochs.

%\vspace{1em}

\subsubsection{Contact safety assessment setup.}
We constructed a regression model using the ResNet50 architecture~\cite{koonce2021resnet}, which takes tactile images as input and predicts the safety coefficient, a metric used to quantify the magnitude of local contact pressure. As described in Supplementary~\ref{ST1}, we collected 2,548 tactile images in SimTac, which were obtained from the contact between different-shaped indents and a GelSight sensor. The image data were split into training, validation, and test sets with a 3:1:1 ratio. Additionally, 559 tactile images were collected from real-world scenarios for Sim2Real transfer evaluation. For each simulated image, we recorded the local peak distributed force in the predicted force field, normalised it to the range of 0 to 1, and used it as the label representing the contact safety coefficient. In the real-world scenario, the tactile image acquisition conditions were identical to those in SimTac, and the aligned data shared the same safety coefficient labels. The regression model was trained with a batch size of 32 for 50 epochs, using the Adam optimiser with a learning rate of 0.001, aiming to minimise the MAE.

\subsection{FEM model for generating ground truth data}\label{ST5}

% Provide a table containing all the parameters used to simulate GelTip
% In the real world, we can easily obtain sparse displacement fields or total force ground truth using tactile sensors with markers and force sensors. However, acquiring ground truth for dense tactile fields, such as force distribution, is much more challenging and remains unsolved. Therefore, we use FEM-based method to generate near-realistic tactile field ground truth.

We first model the flexible membrane of the sensor along with the objects it contacts. The membrane is divided into several regions: the outer surface, the inner surface, and the active deformation surface. The outer surface is the area that interacts with the environment and where the primary deformations occur. The inner surface connects to the internal rigid structure and serves as the boundary condition region. The active deformation surface is relevant for membranes capable of active motion, such as those used in translational or rolling sensors. 

We then define the material properties for the simulation as follows: contacting objects are modelled as linear elastic materials, characterised by Young's modulus \( E \) and Poisson's ratio \( v \). The sensor membrane is modelled as a superelastic material using the Neo-Hookean formulation, with parameters \( C_{10} = \frac{\mu}{2} \) and \( D_1 = \frac{2}{\lambda} \), where \( \mu \) and \( \lambda \) denote the shear modulus and the first Lamé parameter, respectively. Under small strains and rotations, the Neo-Hookean model reduces to linear elasticity, with \( \mu \) and \( \lambda \) related to \( E \) and \( v \) by:

\begin{equation}
\mu=\dfrac{E}{2(1+\nu)}, \quad \lambda=\dfrac{E}{3(1-2\nu)}
\end{equation}

% Contact model and boundary conditions
Finally, we configure the contact and loading conditions. The normal contact between the membrane and the object is modelled as hard contact, utilising a normal-Lagrange contact formulation to mitigate penetration effects. In the tangential direction, a penalty-based friction contact model is employed, allowing for elastic sliding when the object sticks to the sensor. A friction coefficient \( \mu_0 \) is defined to regulate the roughness of the contact interface surface.  
For the rigidly supported areas of the membrane, boundary conditions are constrained to limit displacement. Loads are applied to the objects or the actively deforming regions of the membrane, the surfaces are driven to move in a specified direction by defining the direction of motion and velocity. Details on the FEM parameters setup can be found in Algorithm~\ref{alg4}.

\begin{algorithm}
\caption{FEM Setup for GelTip Sensor}
\label{alg4}
\begin{algorithmic}[1]
    \State \textbf{Create File} 
    \State \quad File$\to$New Model Dataset$\to$With Standard/Explicit Model
    
    \State \textbf{Part} 
    \State \quad File$\to$Import$\to$Part$\to$Import GelTip membrane and indenter
    \State \quad Tools$\to$Reference Point$\to$Select the indenter center point
    \State \textbf{Property} 
    
        \State \quad \textbf{if} Sensor Membrane
            \State \quad \quad Set Hyperelastic Material
            \State \quad \quad \quad Density$\to$$Mass\ Density=0.001$
            \State \quad \quad \quad Hyperelastic$\to$Neo Hooke$\to$Coefficients$\to$$C10=0.025, D1=4.1379$
        \State \quad \textbf{else if} Indenter
            \State \quad \quad Set elastic Material
            \State \quad \quad \quad Density$\to$$Mass\ Density=0.008$
            \State \quad \quad \quad Elastic$\to$$Young's\ Modulus=210, Poisson's\ Ratio=0.3$

    \State \textbf{Assembly} 
    \State \quad Create instance: Import membrane and indenter$\to$Set initial pose
    \State \quad Create set: Boundary Surface$\to$inner support layer, Contact Surface$\to$outer coating layer, Reference Point$\to$indenter centre point

    \State \textbf{Step} 
    \State \quad Create Step: Static, General$\to$$Timeperiol=1$$\to$turn on Nlgeom
    \State \quad Create Field Output: $Domain=Contact\ Surface$$\to$$Frequency=0.025s$$\to$$Ouput=UT,NFORC$

    \State \textbf{Interaction}
    \State \quad Create Interaction Property:
    \State \quad \quad $Normal\ Behavior=``Hard Contact"$
    \State \quad \quad$Tangential\ Behavior=Penalty, Friction Coeff=0.3$
    \State \quad \textbf{if} Indenter
        \State \quad \quad Create Constraint$\to$Coupling
        
    \State \textbf{Load}
    \State \quad Create Boundary Condition:
    \State \quad \quad Fix$\to$Sensor Boundary Surface$\to$Type=Symmetry/Antisymmetry/Encastre
    \State \quad \quad Set Load$\to$Indenter$\to$Type=Displacement$\to$Normal/Shear, interval=0.1 mm 

    \State \textbf{Mesh}
    \State \quad \textbf{if} Indenter
    \State \quad \quad $Approximate\ Global\ Size=0.8$$\to$$Element\ Shape=Tet$
    \State \quad \textbf{else if} Sensor Membrane
    \State \quad \quad $Approximate\ Global\ Size=0.4$$\to$$Element\ Shape=Hex(Structured)$

    \State \textbf{Job}
    \State \quad Generate ODB files

    \State \textbf{Post-Processing}
    \State \quad Run script to extract the displacement/force field from ODB file

    \State \textbf{End Algorithm}
\end{algorithmic}
\end{algorithm}
% TODO: A table including FEM parameters

\begin{figure}[ht!]%
\centering
\includegraphics[width=1\textwidth]{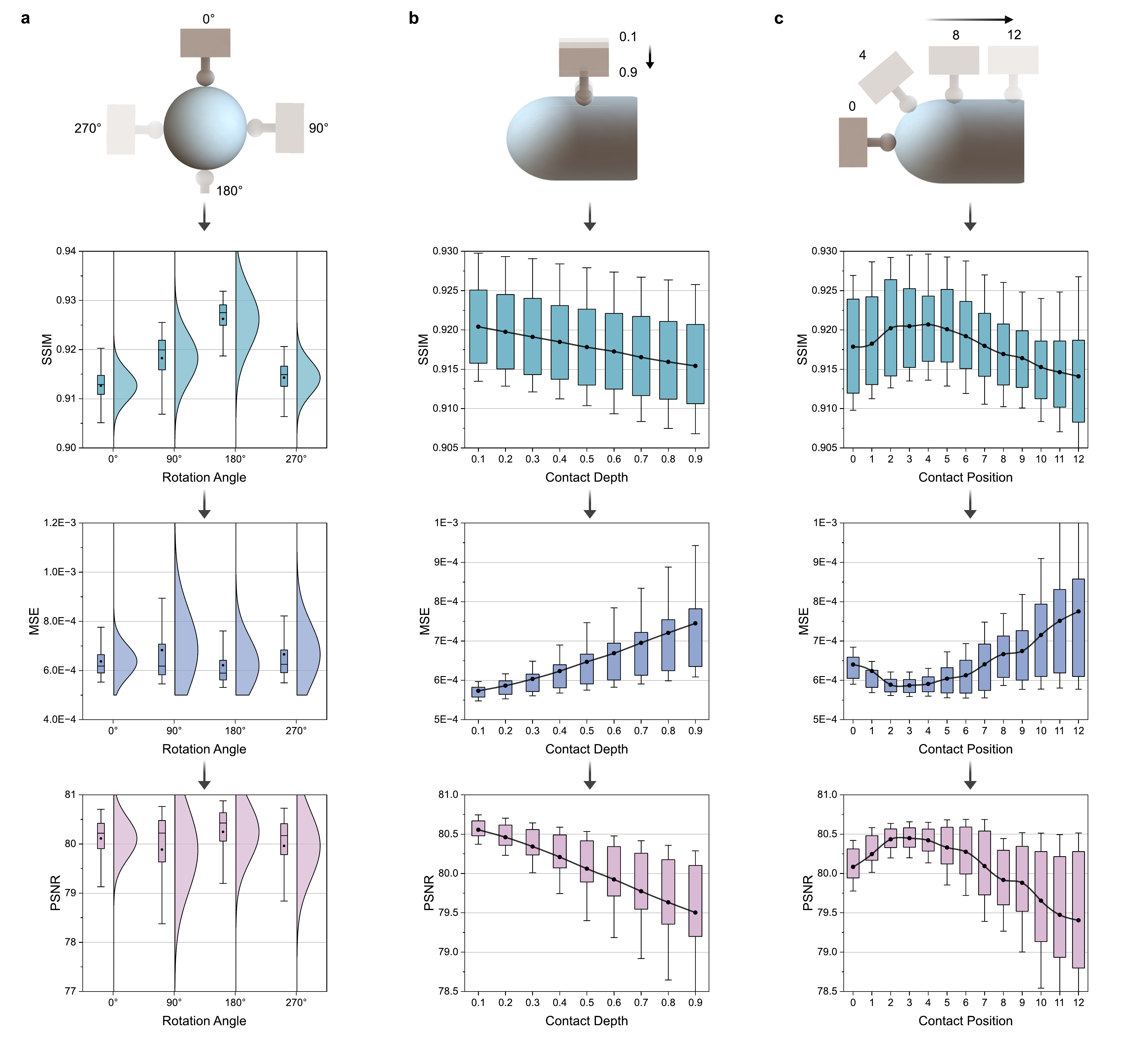}
\caption{Evaluation of optical response simulation under different contact scenarios. The figure presents different contact scenarios (\textbf{a}, different contact angles around the sensor; \textbf{b}, different contact depths; \textbf{c}, different contact positions) and corresponding metrics analysis using SSIM, MSE, and PSNR.  
The similarity between simulated and real images exhibits slight fluctuations at different contact angles and decreases with increasing compression depth. This trend arises from variations in RGB light distribution across contact positions and the expansion of the deformation area at higher compression depths, which leads to greater rendering errors. Additionally, as the contact position shifts from the tip to the base of the sensor, the similarity initially increases before decreasing. This trend arises because the region near the tip extends beyond the camera's focal range, resulting in blurred contact areas in real tactile images. Conversely, in regions closer to the base, the proximity to the camera introduces significant lens distortion, leading to sparse particle distributions and increased interpolation errors in simulated tactile images.}
\label{fig12}
\end{figure}

\begin{figure}[ht!]%
\centering
\includegraphics[width=1\textwidth]{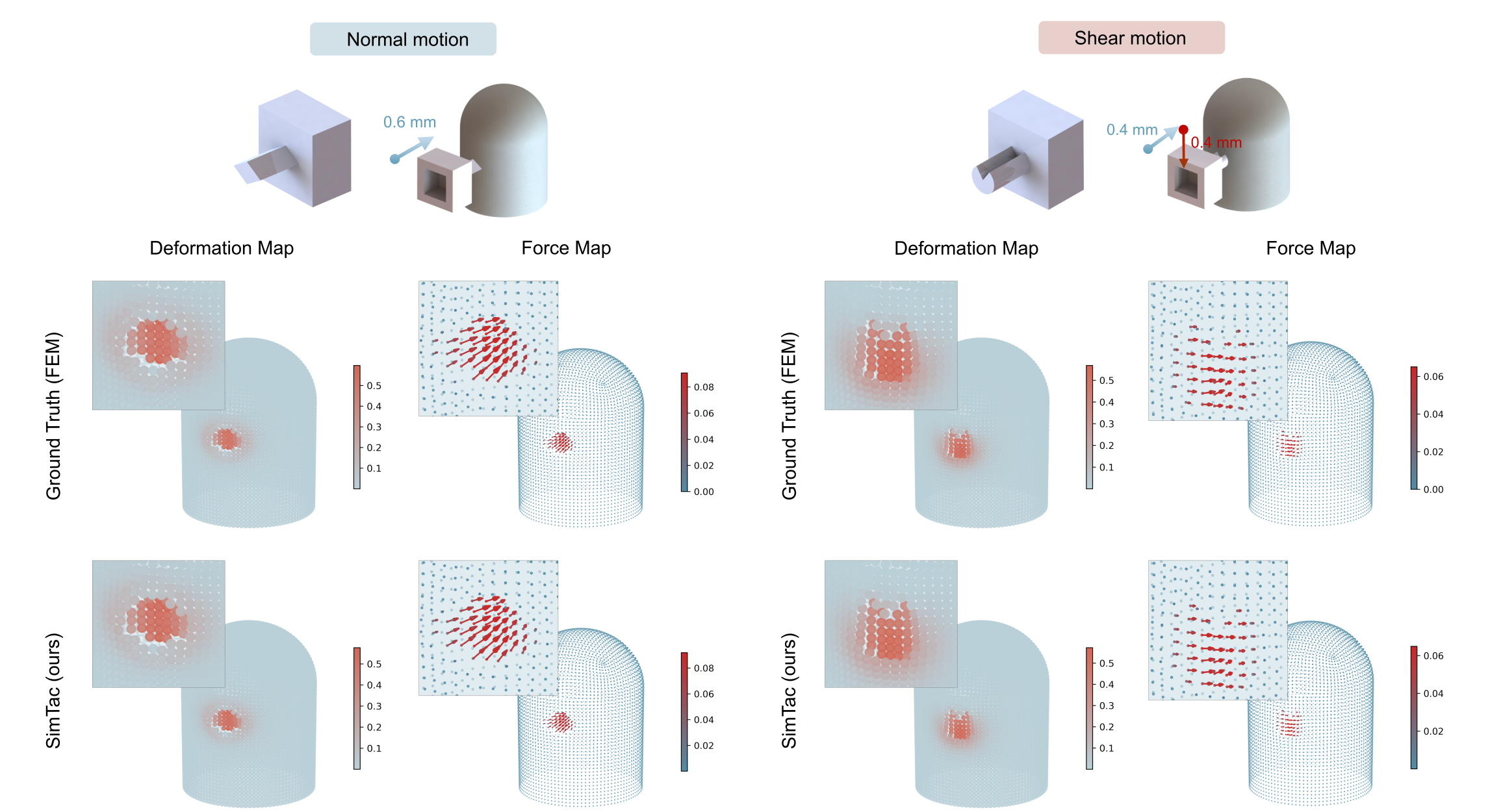}
\caption{Evaluation of mechanical response simulation on the test set. The model was trained on data collected from 10 seen objects and evaluated on 4 unseen objects to assess its generalisation capability. 3D heatmaps were employed to visualise the distributions of deformation and force fields, where the colour represents the magnitude of the field data. The comparison shows that the simulated deformation and force values from the proposed model exhibit particle-level consistency with the ground truth computed from FEM, demonstrating the model’s accuracy in simulating deformation and force when the sensor interacts with objects of various shapes.}
\label{fig13}
\end{figure}

\begin{figure}[ht!]%
\centering
\includegraphics[width=1\textwidth]{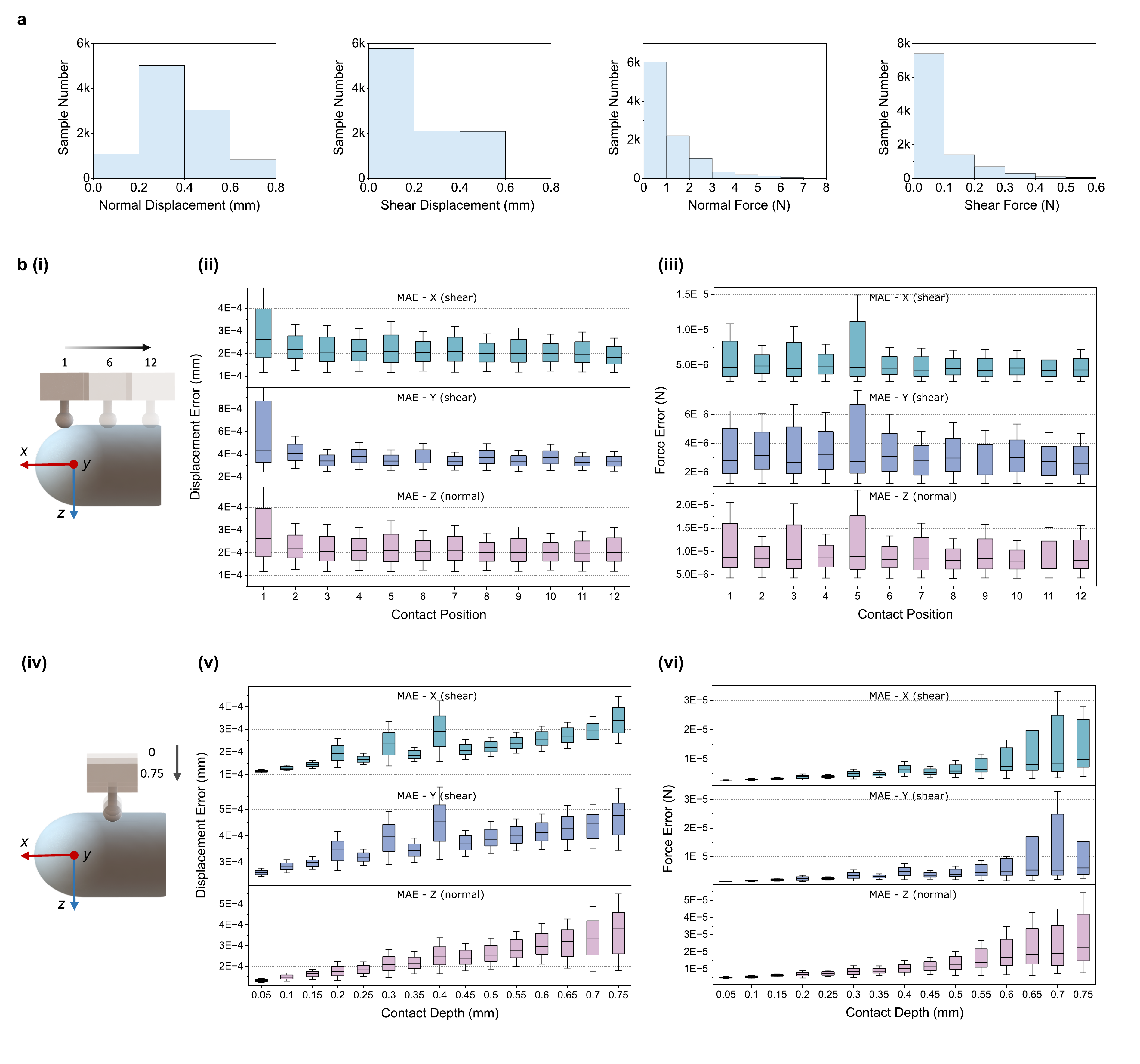}
\caption{Evaluation of mechanical response simulation under different contact scenarios. \textbf{a}, Histogram of the tactile image quantity distribution across varying contact displacements and forces. \textbf{b}, Analysis of MAE between the simulated mechanical response and ground truth (computed by FEM) for the dense deformation and force fields at various (\textbf{i})-(\textbf{iii}) contact positions and (\textbf{iv})-(\textbf{vi}) depths. The deformation and force prediction errors are notably larger in contact region 1, which represents the transition zone between the sensor tip and its cylindrical body. This region experiences insufficient contact between the object and the sensor, resulting in substantial prediction errors. Additionally, the prediction errors increase with compression depth, as both displacement and total force values rise, thereby amplifying the overall prediction inaccuracies.}
\label{fig14}
\end{figure}

\begin{figure}[ht!]%
\centering
\includegraphics[width=1\textwidth]{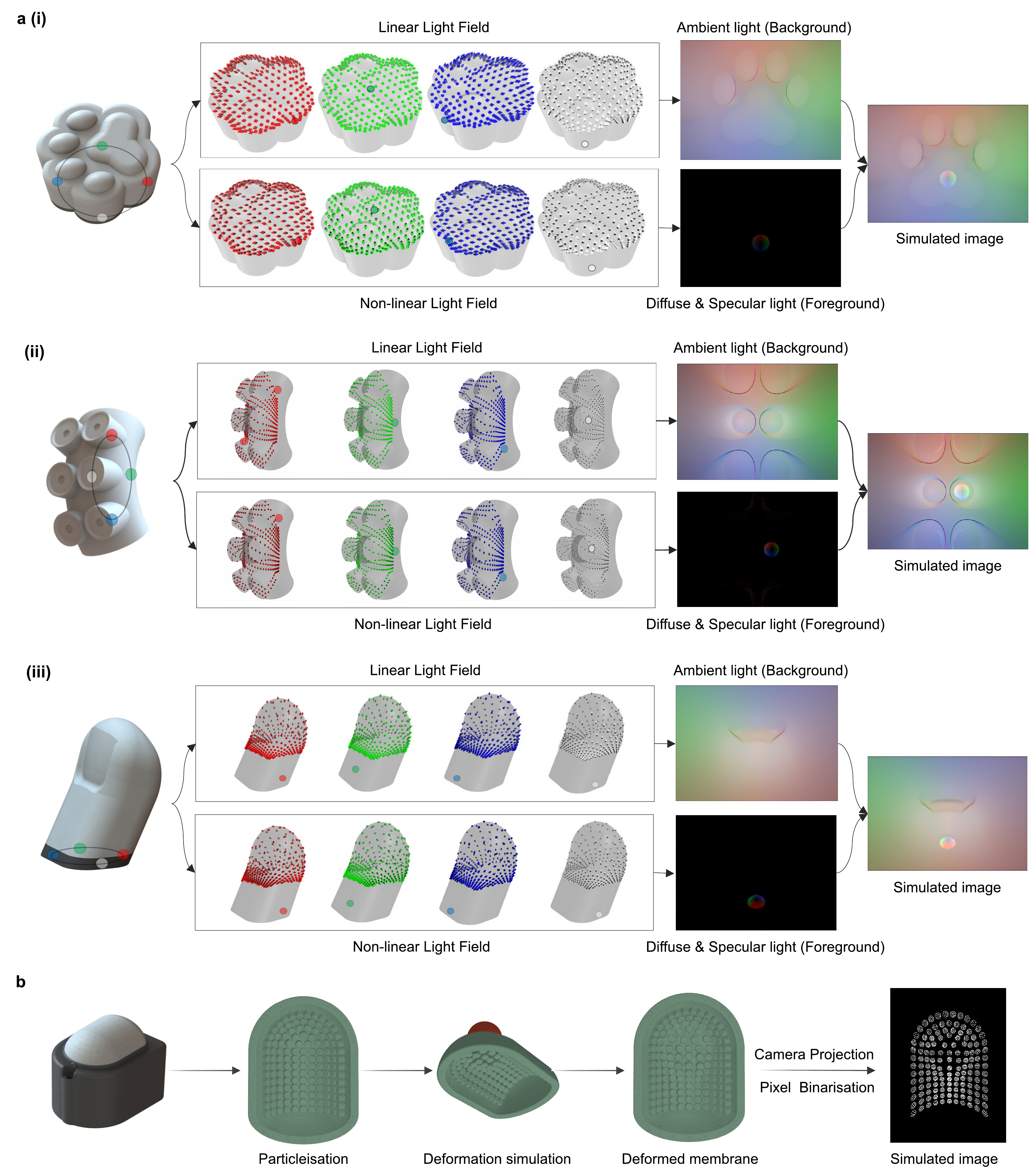}
\caption{Simulation process for optical responses of biomorphic vision-based tactile sensors with diverse sensor shapes and optical configurations. \textbf{a}, Simulation process for optical responses of markerless GelSight-type tactile sensors: inspired from (\textbf{i}) cat paws, (\textbf{ii}) octopus tentacles, and (\textbf{iii}) human thumb. We reduced the density of the light field to facilitate visualisation and understanding. The tactile image is composed of a background image calculated from a linear light field and a foreground image calculated from a nonlinear light field. \textbf{b}, Simulation process for optical responses of marker-based TacTip-type tactile sensors: DigiTac sensor~\cite{lepora2022digitac}. The simulator can replicate the deformation and motion of TacTip's physical pins during contact. The marker-based tactile images are generated by tracking the positions of particles within the tip region of the pins.}
\label{fig15}
\end{figure}

\begin{figure}[ht!]%
\centering
\includegraphics[width=1\textwidth]{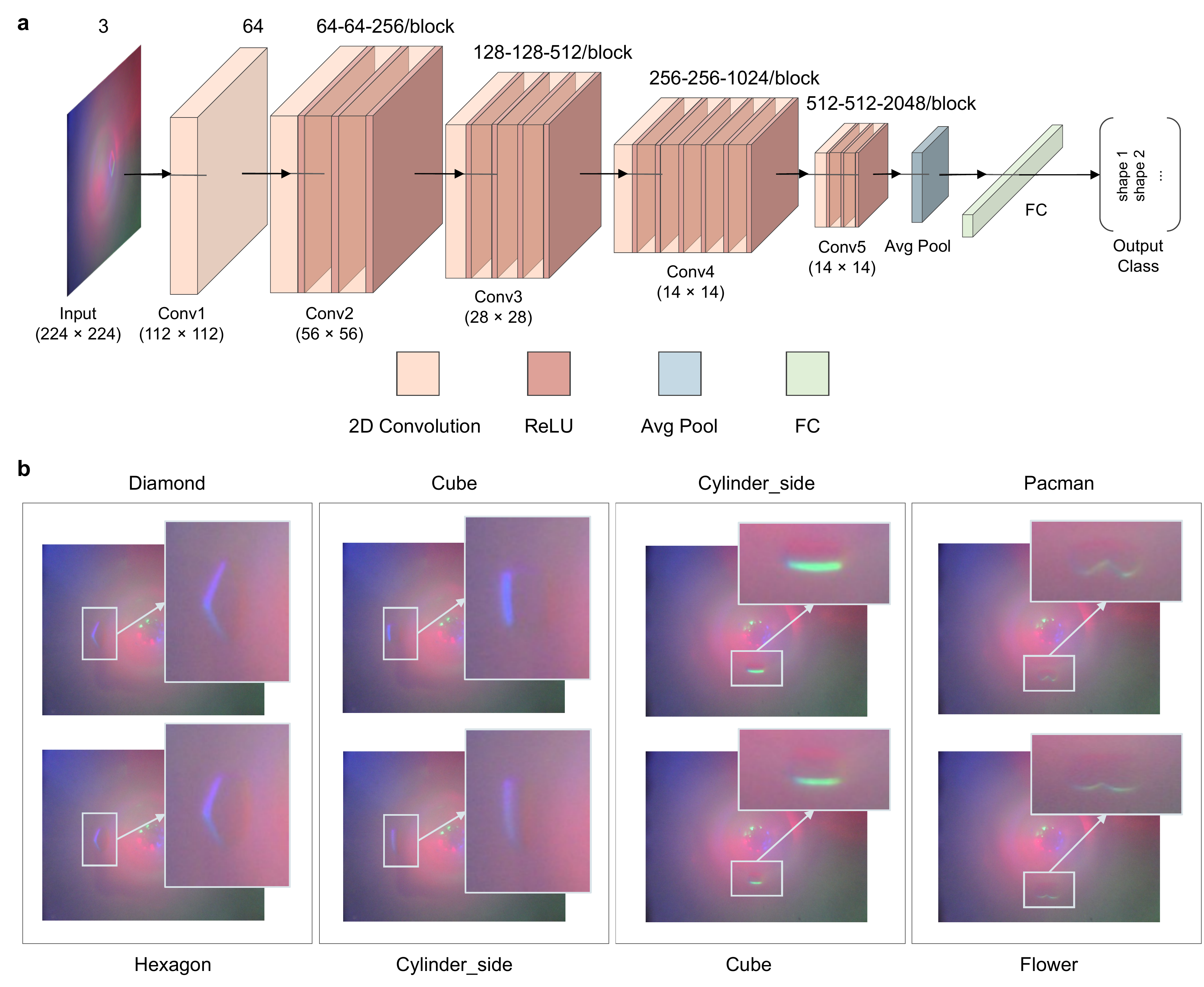}
\caption{Network structure in Sim2Real object classification task and similar shapes leading to incorrect predictions. \textbf{a}, The ResNet50~\cite{koonce2021resnet} architecture is employed as the backbone for the proposed classification model. \textbf{b}, Example of similar shapes leading to incorrect predictions.}
\label{fig16}
\end{figure}

\begin{figure}[ht!]%
\centering
\includegraphics[width=1\textwidth]{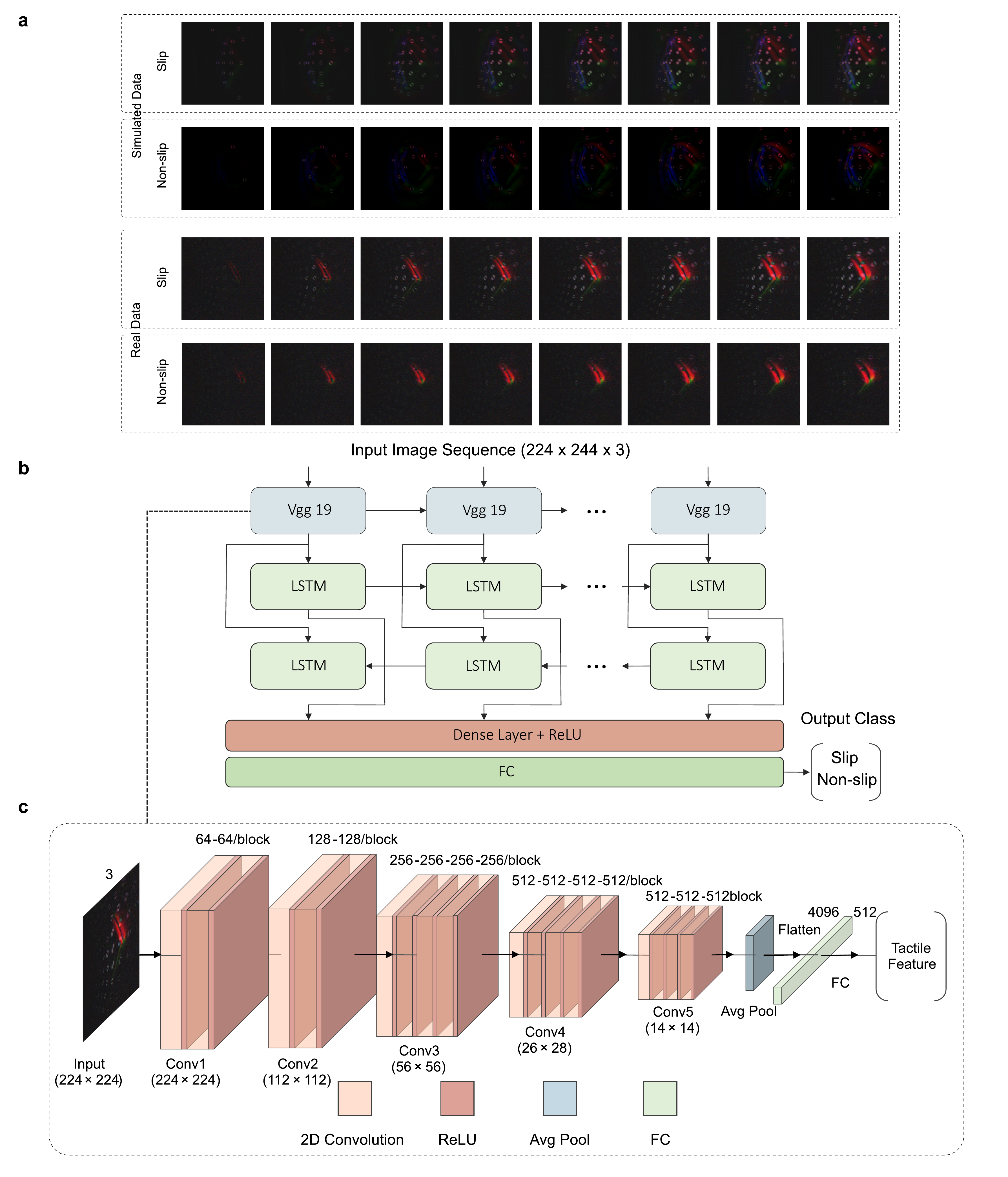}
\caption{Network structure in Sim2Real slip detection task. \textbf{a}, Examples of simulated and real-world tactile image sequences as input to the network. In non-slip cases, the object’s contour and surface markers move synchronously, while in slip cases, the object continues to move as the markers remain nearly stationary, indicating relative slip motion. \textbf{b}, VGG-19~\cite{tammina2019transfer} architecture is employed for tactile feature extraction, and LSTM model~\cite{graves2012long} is utilized for slip and non-slip prediction. \textbf{c}, Details of the employed VGG-19 architecture.}
\label{fig17}
\end{figure}

\begin{figure}[ht!]%
\centering
\includegraphics[width=1\textwidth]{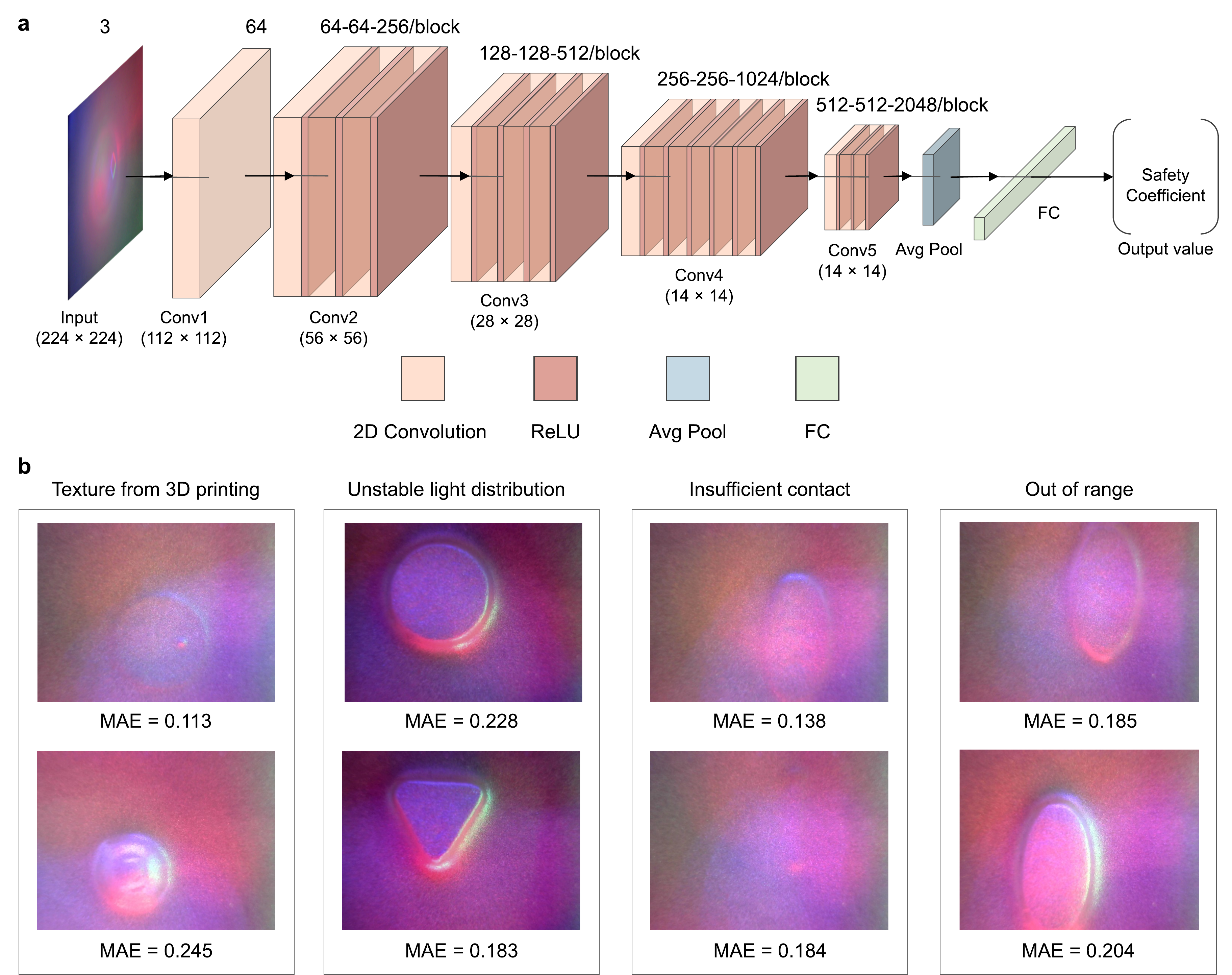}
\caption{Network structure in Sim2Real contact safety assessment task and failure cases with large prediction errors. \textbf{a}, The ResNet50~\cite{koonce2021resnet} architecture is employed as the backbone for the proposed regression model. \textbf{b}, Examples of failure cases with large prediction errors in the safety coefficient. The higher Sim2Real error compared to Sim2Sim can be attributed to several factors: Firstly, 3D-printed objects exhibit surface textures absent in simulations, introducing artefacts that affect prediction error. Secondly, as the indentation depth increases, the deformation of the silicone surface alters light distribution, causing the overall image to darken. Thirdly, some real-world data suffer from insufficient indentation, resulting in tactile images that only capture partial contact contours rather than the full contact area. Finally, certain contact poses may cause objects to extend beyond the camera’s field of view, limiting the completeness of tactile information.}
\label{fig18}
\end{figure}

\end{document}